\documentclass[journal]{IEEEtran}

\ifCLASSINFOpdf

\else

\fi

\usepackage{hyperref}
\usepackage{subfigure}
\usepackage{makeidx}  
\usepackage{graphicx}
\usepackage{amsmath}
\usepackage{nccmath}
\usepackage{extarrows}
\usepackage{amsfonts,amsmath}
\usepackage{algorithm2e}
\usepackage{changepage}
\usepackage{float}

\usepackage{ragged2e}
\usepackage{cite}

\begin{document}

\title{Real-time Dense Reconstruction of\\Tissue Surface from Stereo Optical Video
}

%

\author{Haoyin Zhou, ~\IEEEmembership{Member,~IEEE}
        and Jayender Jagadeesan,~\IEEEmembership{Member,~IEEE}
\IEEEcompsocitemizethanks{\IEEEcompsocthanksitem
 Haoyin Zhou and Jayender Jagadeesan are with the Surgical Planning Laboratory, Brigham and Women's Hospital, Harvard Medical School, Boston,
 MA, 02115, USA.  Jayender Jagadeesan owns equity in Navigation Sciences, Inc.  He is a co-inventor of a navigation device to assist surgeons in tumor excision that is licensed to Navigation Sciences.  Dr. Jagadeesan's interests were reviewed and are managed by BWH and Partners HealthCare in accordance with their conflict of interest policies.\protect\\
E-mail: zhouhaoyin@bwh.harvard.edu; jayender@bwh.harvard.edu.
}

\thanks{}}

\markboth{}%
{Shell \MakeLowercase{\textit{et al.}}: Bare Demo of IEEEtran.cls for Computer Society Journals}

\IEEEcompsoctitleabstractindextext{%
\begin{abstract}
\justifying
We propose an approach to reconstruct dense three-dimensional (3D) model of tissue surface from stereo optical videos in real-time, the basic idea of which is to first extract 3D information from video frames by using stereo matching, and then to mosaic the reconstructed 3D models. To handle the common low texture regions on tissue surfaces, we propose effective post-processing steps for the local stereo matching method to enlarge the radius of constraint, which include outliers removal, hole filling and smoothing. Since the tissue models obtained by stereo matching are limited to the field of view of the imaging modality, we propose a model mosaicking method by using a novel feature-based simultaneously localization and mapping (SLAM) method to align the models. Low texture regions and the varying illumination condition may lead to a large percentage of feature matching outliers. To solve this problem, we propose several algorithms to improve the robustness of SLAM, which mainly include (1) a histogram voting-based method to roughly select possible inliers from the feature matching results, (2) a novel 1-point RANSAC-based P$n$P algorithm called as DynamicR1PP$n$P to track the camera motion and (3) a GPU-based iterative closest points (ICP) and bundle adjustment (BA) method to refine the camera motion estimation results. Experimental results on \textit{ex-} and \textit{in vivo} data showed that the reconstructed 3D models have high resolution texture with an accuracy error of less than 2 mm. Most algorithms are highly parallelized for GPU computation, and the average runtime for processing one key frame is 76.3 ms on stereo images with $960\times 540$ resolution.

\end{abstract}

\begin{IEEEkeywords}
Surface Reconstruction; Stereo Matching; SLAM; GPU Parallel Computation; Stereo Imaging
\end{IEEEkeywords}}

\maketitle

%
\IEEEdisplaynotcompsoctitleabstractindextext

\IEEEpeerreviewmaketitle

\section{Introduction}

\IEEEPARstart{T}{he} surgeon's visualization during surgery is typically limited to the anatomical tissue surface exposed to him/her through an optical imaging modality, such as a laparoscope, endoscope or microscope. As a result, intraoperative identification of the critical structures lying below the visual surface is difficult and could lead to inadvertent complications during the surgery. To solve this problem, many surgical navigation systems utilize models of tissue surface, internal structures and tumors segmented from preoperative MR/CT imaging for intraoperative guidance. However, direct registration between two-dimensional (2D) optical (microscopy, endoscopy or laparoscopy) videos and three-dimensional (3D) MR/CT images is difficult and highly non-trivial. To overcome the difficulty in registering the multimodal images, 3D information can be extracted from 2D optical videos, which is still an open problem and is especially challenging when the surface texture is low. In this paper, we propose a series of novel methods to reconstruct textured 3D models of tissue surfaces from stereo optical videos in real-time. The textures on the reconstructed tissue surface models have the same resolution as the input video frames, which can greatly facilitate surgical navigation for the following reasons: (1) During surgery, only a small area of the target tissue may be exposed and landmarks that can be automatically recognized are often invisible. In addition, blood or surgical smoke may occlude the target tissue. Hence, it is important to provide high resolution textures to help the clinicians to recognize the tissue from the reconstructed models and then perform manual registration. (2) Intuitive visual feedback as part of a surgical navigation system is also very important for tumor localization. And with high resolution textures, clinicians are able to visualize the \textit{in vivo} scene from different angles intuitively.

Stereo optical imaging modalities have been widely used in the operating room to provide depth perception to the surgeon. In the past decade, many efficient stereo matching methods have been proposed to estimate depths of image pixels by establishing pixel-to-pixel correspondences between the stereo images, the results of which can be further refined to generate fine 3D models. Stereo matching methods can be roughly classified into global and local methods. Global methods use constraints on scan-lines or the whole image \cite{yang2015stereo}\cite{hirschmuller2008stereo}, which are able to handle low texture regions by using explicit or implicit interpolation. However, global methods have high computational complexity and are inappropriate for real-time applications. In contrast, local methods only use constraints on a small number of pixels surrounding the pixel of interest \cite{bleyer2011patchmatch}, which are fast but are difficult to handle low texture regions. In this paper, we propose effective outliers removal, hole filling and smoothing methods as the post-processing steps for the local stereo matching methods, which have low computational complexity low and are appropriate for graphics processing unit (GPU) parallel computation.

Stereo matching-based 3D reconstruction is highly dependent on the texture of the observed object. However, the surface texture of tissues, such as lung and liver, is not rich enough to be observed at a distance due to the limited camera resolution and poor illumination condition. Another important reason to use a small camera-tissue distance is that the baseline of stereo imaging modalities is usually short, which result in large uncertainties when estimating large depths. However, due to the limited field of view, a small camera-tissue distance will lead to only a small area of the surface that can be reconstructed from the pair of stereo images, which is insufficient to perform accurate registration between pre- and intraoperative 3D models \cite{mountney2010three}. To solve the contradiction between the accuracy of 3D reconstruction and registration, we propose to scan the tissue surface at a close distance and perform stereo matching on the acquired stereo images, then mosaick the 3D models at different time steps according to model alignment obtained by simultaneously localization and mapping (SLAM).

SLAM is one of the most important topics in the robotics navigation field, which aims to estimate the camera motion and reconstruct the surrounding environment in real-time \cite{mur2015orb}\cite{mur2017orb}. To date, SLAM methods have proven effective in reconstructing large environments and estimating long motions \cite{engel2014lsd}, hence it is a reasonable assumption that the accumulative errors of SLAM methods is minimal for the small \textit{in vivo} environments. SLAM methods are often based on feature points matching to establish correspondences between video frames. However, for tissue surfaces with low and/or repeating texture under varying illumination conditions, feature matching is challenging \cite{puerto2013fast} and a large percentage of matching outliers may cause failure of the SLAM methods. In order to overcome the difficulties in feature matching and improve the robustness of mosaicking, we first propose a novel histogram voting-based method to select possible inliers from the feature matching results. Then, using the selected possible inliers as the control points, we extend our previous work \cite{zhou2018re} and propose a novel perspective-$n$-points (P$n$P) algorithm called as DynamicR1PP$n$P to estimate the camera motion, which can remove incorrect and build new matches dynamically. Finally, we propose to integrate feature matching and iterative closest points (ICP)-based costs into an optimization method to refine the camera motion estimation results. The main algorithms involved in our SLAM framework are implemented in CUDA C++ and run on the GPU.

This paper is organized as follows: In Section II, we describe the process of the stereo matching method and provide the details of its GPU implementation. The SLAM-based model mosaicking method, including histogram voting-based inliers selection, DynamicR1PP$n$P and GPU-based BA+ICP, is introduced in Section III. Evaluation results on \textit{ex vivo} and \textit{in vivo} data are presented in Section IV. A discussion is presented in Section V.

\subsection{Related Works}

Optical video-based 3D reconstruction of tissue surfaces can improve the accuracy of intraoperative navigation \cite{maier2013optical}. Stereo matching is one of the most effective 3D reconstruction methods in surgical navigation applications \cite{hirschmuller2009evaluation}, which estimates pixel disparities by comparing stereo images. Stereo matching is an important and active topic in the computer vision field and a large number of effective methods exist, and reader may refer to the Middlebury website \cite{scharstein2002taxonomy} for the list of stereo matching methods. Stereo matching methods can be roughly classified into global and local methods. Global stereo matching \cite{scharstein2002taxonomy}, such as dynamic programming \cite{birchfield1999depth} and graph cuts \cite{boykov2001fast}, exploit nonlocal constraints to reduce sensitivity to regions that fail to match due to low texture and occlusions, which make explicit smoothness assumptions to solve an optimization problem. However, the high time complexity makes global stereo matching difficult to be real-time \cite{wang2008region}, hence most current real-time 3D reconstruction systems are based on local stereo matching.

Local stereo matching methods estimate disparities of pixels by computing matching matrices between small and local image patches. There exist many metrics to evaluate the similarity between two image patches \cite{brown2003advances}. The most straightforward one is window-based matching costs, which compare the differences of squared image windows. Zero-mean normalized cross-correlation (ZNCC) \cite{di2005zncc} is one of the most effective window-based costs due to its good robustness to illumination changes. However, such squared window-based methods cannot handle pixels near object edges because they may belong to different surfaces. To overcome this problem, non-parametric matching costs, such as rank and census methods\cite{zabih1994non} and ordinal measures \cite{bhat1998ordinal}, were proposed to handle object boundaries. Another class of effective methods is based on support window methods \cite{zhang2009cross}, such as PatchMatch \cite{bleyer2011patchmatch}, which uses varying shape of the matching window. To achieve better accuracy,  researchers propose to dynamically update the weights of pixels within the support window \cite{yoon2006adaptive}. For our task, the needs of handling tissue edges or occlusion are not high because usually only one target tissue needs to be reconstructed and the surgeons may simply remove the instrument during the scan. Hence we use ZNCC matching in our method, which is fast on the GPU. Our main contribution of the stereo matching part is that we propose several effective post-processing steps to address the low texture problem, which can also be used for the refinement of other local stereo matching methods.

 Many real-time stereo matching systems are based on ZNCC \cite{di2005zncc}. To achieve real-time performance, it is essential to reduce the number of candidate disparities for local stereo matching methods. For example, Bleyer et al\cite{bleyer2011patchmatch} proposed an effective disparities searching strategy by first generating disparities for all pixels randomly, and then iteratively replacing the disparity of a pixel with that of its neighboring pixel if the new value suggests a better ZNCC matching. Stoyanov et al \cite{stoyanov2010real}\cite{chang2013real} matched a sparse set of salient regions using stereo Lucas-Kanade and propagated the disparity around each matched region. They reported a 10Hz updating rate for images with $360\times 240$ resolution. The development of GPU or FPGA \cite{jin2010fpga} based parallel computational algorithms can greatly accelerate the image patch matching process \cite{rohl2012dense}. Zollh{\"o}fer et al \cite{zollhofer2014real} reported a 100 Hz update rate for stereo images with $1280\times 1024$ resolution using a NVIDIA Titan X GPU. Our CUDA C++ implementation achieves a 200 Hz updating rate for the $960\times 540$ resolution and 100 candidate disparities, which is sufficient for our surface reconstruction system.

3D models generated by stereo matching are limited to the field of view, which may be too small for surgical guidance. Structure-from-motion (SfM) \cite{sun2013surface} or simultaneously localization and mapping (SLAM)\cite{grasa2014visual}\cite{mountney2010motion}\cite{chen2018slam} methods are able to align video frames at different time steps and generate a much larger synthetic field of view, which have been employed for 3D reconstruction of tissues. For example, Mountney et al \cite{mountney2009dynamic} proposed to expand the field of view based on SLAM. Most SfM and SLAM methods only reconstruct sparse feature points, which poorly describe the surgical scene.

Dense SLAM methods have also been developed to generate dense tissue models in real-time. Totz et al \cite{totz2011dense} proposed an EKF-SLAM-based method for dense reconstruction. EKF-SLAM suffers from low accuracy and is difficult for representing loop closing. Recently, Mahmoud et al \cite{mahmoud2018live} proposed a monocular vision-based dense tissue 3D reconstruction method by using ORB-SLAM \cite{mur2017orb} to estimate the camera motion. However, because ORB-SLAM is based on ORB features and RANSAC+P3P\cite{lepetit2009epnp} for camera motion tracking and loop closing, its robustness is not satisfying with low texture scenes. In this paper, we propose novel camera motion tracking algorithms and a more robust SLAM framework to improve the robustness of camera pose estimation with low texture surfaces.

Another effective way to perform real-time dense reconstruction is to combine sparse SLAM and stereo vision, the idea of which is closely related to the famous KinectFusion work \cite{newcombe2011kinectfusion}, which merges the raw depth map provided by Microsoft Kinect to generate the fine models. It is a natural idea to replace the depth map with the results of stereo matching. However, the most difficult part is to align the depth map by SLAM, and
KinectFusion is based on the ICP method. However, due to the narrow field of view and the smooth surface of tissue, ICP-based alignment cannot achieve accurate registration in the tangential directions.

\section{Stereo Matching}

\begin{figure*} [ht]
\vspace{0.0cm}
\centering
 \subfigure[]{
 \includegraphics[height=4.5cm]{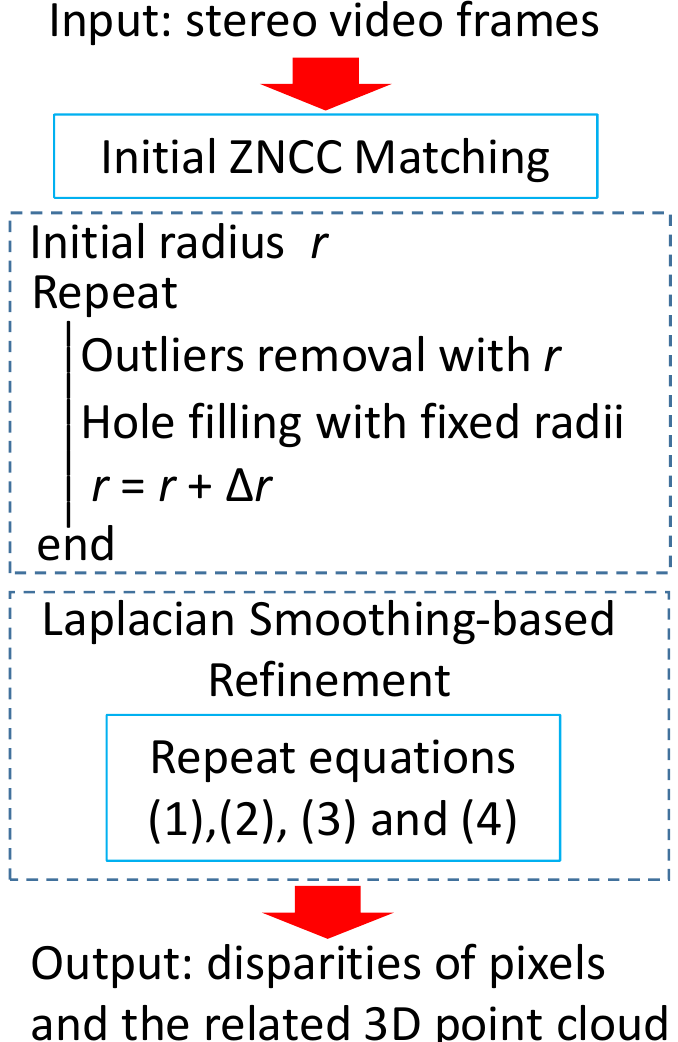}
}
 \subfigure[]{
 \includegraphics[height=4.5cm]{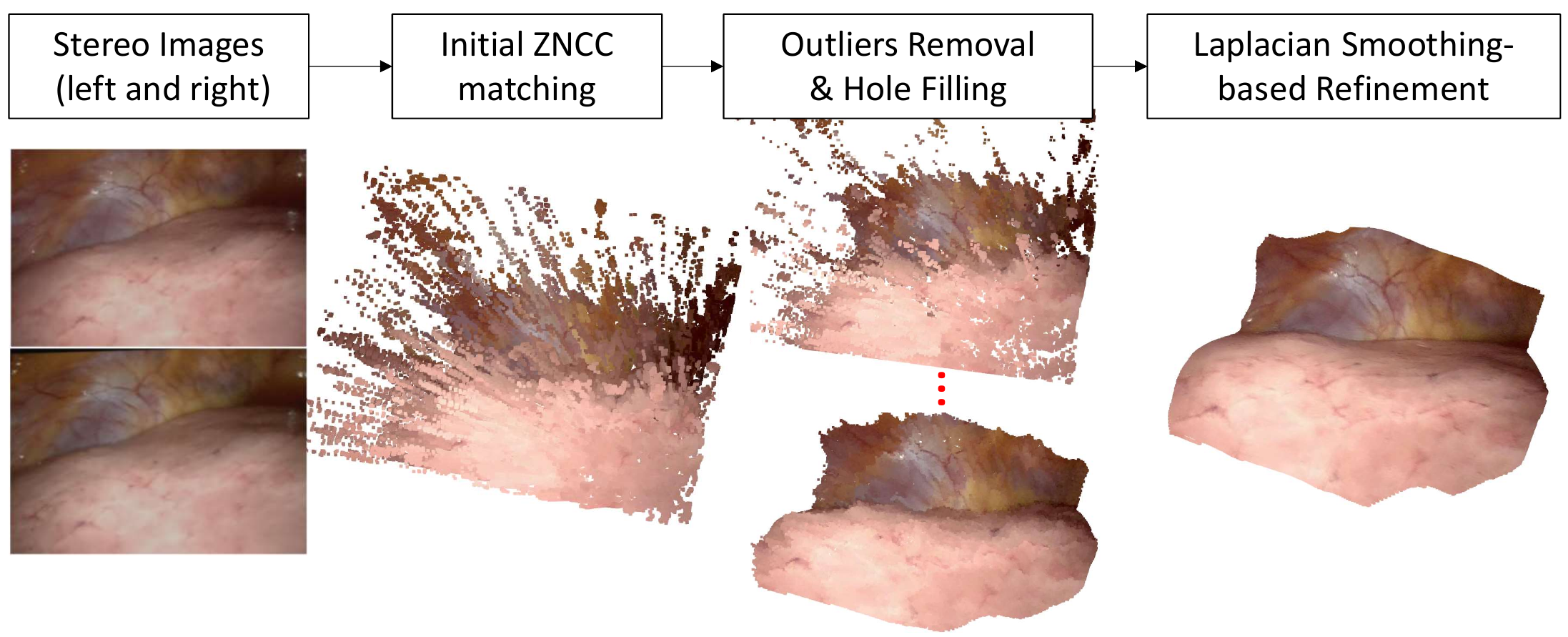}
}
 \caption{(a) The flow chart of our stereo matching method. (b) An intuitive example to show the stereo matching process with a pair of stereo laparoscopic images captured during a lung surgery at our hospital, the texture on the tissue surface is low.}
\label{fig_BinocularProcess}
\end{figure*}

After stereo camera calibration, physical depthes of stereo image pixels can be directly computed from the disparities. We used the Matlab Computer Vision Toolbox to calibrate the stereo laparoscope and our C++ code to convert image disparities to physical depthes is equivalent to the Matlab 'reconstructScene' function. For local stereo matching methods, the estimation of disparities at low texture regions is difficult due to the lack of direct corresponding information between left and right images. However, low texture regions are common on tissue surfaces due to tissue optical properties, limited image resolution, poor image quality and poor illumination conditions. Most stereo matching methods rely on interpolation to propagate information from highly textured regions to low texture regions. For example, by interpolating between edges, a textureless flat wall can be reconstructed accurately. However, tissue surfaces have more complex shapes, and interpolation-based methods may not be accurate at distant regions. Hence, we do not seek to estimate disparities of all pixels in the stereo matching step, but rely on the subsequent mosaicking step to generate more complete and larger models of tissue surfaces.

To overcome the high time complexity drawback of global stereo matching methods and difficulty to handle low texture regions of local stereo matching methods, we propose a novel stereo matching framework as shown in Fig. \ref{fig_BinocularProcess} to enlarge the radius of constraints of local stereo matching. First, we employed the zero-mean normalized cross correlation (ZNCC) metric to evaluate similarities between local image patches to estimate disparities of pixels. Then, we developed a robust outliers removal and hole filling method to refine the ZNCC matching results. The first two steps provide discrete initial disparity values that are from the candidate disparities pool for the final refinement step, where we integrate the ZNCC metrics and the smoothing cost into a modified Laplacian smoothing framework. This method is able to build large connections among pixels when minimizing the cost function, and is easier to compute than conventional Gauss-Newton (GN) or Levenberg-Marquardt (LM) algorithms. It is worth clarifying that we are not implying that ZNCC is the best metrics, however since our stere matching methods are mostly post-processing steps, it is easy to replace ZNCC with other local matching metrics. The algorithms in our stereo matching method work in parallel with respect to each pixel, and are highly appropriate for GPU parallel computing.

\subsection{ZNCC-based Local Matching}

The most widely used local stereo matching method first generates disparities for all pixels randomly, and then iteratively replaces the disparity of a pixel with that of its neighboring pixel if the new value suggests a better matching \cite{bleyer2011patchmatch}. This process has demonstrated high efficiency and even CPU-based serial computation can be real-time (2-3 Hz). Another advantage is that this type of method implicitly takes into account smoothing among pixels. However, in practice we found that this method is not suitable for the case of smooth tissue surface because pixels that have the same disparity are often distributed in a narrow belt, which makes it difficult to propagate a correct disparity value and many iterations are needed. In addition, these methods cannot make full use of the GPU parallel computing ability, because the propagation process can only be parallelized to $W$ and $H$ threads alternatively, where $W$ and $H$ are image width and height respectively.

Our stereo matching method is based on the ZNCC metrics to evaluate similarities between local image patches. In our experiments we use a window size of $11\times 11$ pixels. To make full use of GPU parallel computing ability, we develop a brute force way by launching GPU threads for each pixel to test the candidate disparity values. To achieve higher computational speed, in the matching window we only use every other pixel values, which is distributed as a chessboard. The details of our GPU implementation are briefly described as follows: For images with a resolution of $W \times H$, our CUDA implementation launches $H$ CUDA blocks and each CUDA block has $W$ threads. We cache neighboring image rows into the GPU shared memory for each CUDA block to avoid the slow I/O speed of global memory. With a $960\times540$ resolution and 100 candidate disparity values, the runtime of our GPU-based ZNCC matching method is around 5ms.

\subsection{Outliers Removal and Hole Filling}

The initial ZNCC matching may result in a large amount of outliers. Our outliers removal and hole filling method is under a reasonable assumption that the tissue surface is relatively smooth. Hence, an inlier should have sufficient number of neighboring points that has smooth change of disparities. Denoting $r$ as the detection radius, we detect along each 8-radial directions within radius $r$ and check if the disparity of two neighboring points is smaller than a pre-defined threshold ($=2.5$ in our experiments). If none of the 8-radial directions satisfies this smooth disparity assumption, the point will be recognized as an outlier and removed.

We developed two hole filling methods. For a left-image pixel that cannot find its corresponding right-image pixel, the first method searches along the pixel's 8-radial directions and the second method searches within a radius of the pixel. In our experiment the two radii for the hole filling methods are fixed, which are 50 and 20 pixels respectively. If sufficient number of neighboring points have a valid disparity value, then disparity of this pixel is filled according to interpolation. In the hole-filling step the iterations are performed within a radius, which avoids interpolation at distant areas.

However, when removing outliers, it is difficult to pre-define a radius $r$ for all cases. A small $r$ may keep too many outliers and a large $r$ may remove inliers. To removal outliers and preserve as many inliers as possible, we propose to use an iterative process that alternately performs outlier removal and hole filling, as shown in Fig. \ref{fig_BinocularProcess}(a). In this process we gradually enlarge $r$ with a step $\Delta r$ when detecting outliers. Hence, disparities that are removed may then be filled, and neighboring inliers will not be removed with larger $r$. In our experiments, the number of outliers removal and hole filling iterations is 3; the radius $r$ is 10 pixels initially and increases at a step of $\Delta r = 10$ pixels.

\subsection{Improved Laplacian Smoothing-based Refinement}

Further step to refine the estimated disparities is necessary because (1) the initial disparities after the first two steps are discrete values that are directly selected or interpolated from the candidate disparities and (2) relationships among pixels are not fully considered. Our refinement method is based on Vollmer's improved Laplacian smoothing method \cite{vollmer1999improved}, which is able to avoid model shrinking compared with standard Laplacian smoothing. We integrate a cost function that consists of the ZNCC metrics and the smoothing cost into this improved Laplacian framework to allow for dynamically updating the disparities. The details of our refinement step are as follows:

We denote the discrete disparity of a pixel $i$ as $o_i$, which initially is equal to the disparity value after the first two steps. The smoothed disparities at the $k$th iteration are denoted as $d_i^{(k)}$. After an initialization $d_i^{(0)} := o_i$, the refinement method performs the following steps in the $k$th iteration:

\begin{equation}
d_i^{(k)}: = {\mathop{\rm average}\nolimits} ({o_j}),
\label{eq_laplacine_1}
\end{equation}
\noindent where $j$ is the index of neighboring pixels of point $i$ within a pre-defined radius, and we use the smoothing radius of 15 pixels in our experiments.

\begin{equation}
b_i := d_i^{(k)} - \alpha o_i - (1 - \alpha) d_i^{(k - 1)},
\label{eq_laplacine_2}
\end{equation}

\noindent where $b_i$ is introduced to avoid model shrinking. $\alpha \in [0,1]$ is a weighting coefficient and $\alpha = 0.1$ in our experiment. And then

\begin{equation}
d_i^{(k)}: = d_i^{(k)} - {\mathop{\rm average}\nolimits} ({b_j}).
\label{eq_laplacine_3}
\end{equation}

Equations \eqref{eq_laplacine_1}, \eqref{eq_laplacine_2} and \eqref{eq_laplacine_3} are derived from Vollmer's Laplacian smoothing method, which generate continuous disparities $d_i$ by smoothing discrete disparities $o_i$. We further propose to update the discrete disparities $o_i$ in each iteration according to the minimization of a cost function that consists of the ZNCC metrics and the smoothing cost. Specifically, with an updated disparity $d_i^{(k)}$ in the iteration, we search within a disparity range $[d_i^{(k)} - 5, d_i^{(k)} + 5]$, and update the $o_i $ to the disparity value that minimizes

\begin{equation}
o_i := \arg \mathop {\min }\limits_{o_i^*} {f_{{\rm{zncc}}}}({o_i^*}) + \eta {f_{{\rm{smooth}}}}({o_i^*} - {d_i}),
\end{equation}

\noindent where ${f_{{\rm{zncc}}}}(o_i^*)$ is the ZNCC matching cost, which equals to the reciprocal of the ZNCC matching value when using a disparity $o_i^*$. ${f_{{\rm{smooth}}}}({o_i^*} - {d_i}) = ({o_i^*} - {d_i})^2$ is the smoothing cost because $d_i$ is the smoothed value of neighboring $o_j$. $\eta$ is a coefficient. The size of matching window affects ${f_{{\rm{zncc}}}}(o_i^*)$, and with a $11\times11$ pixels window, we use $\eta = 0.01$ in our experiments.

The advantage of using this improved Laplacian smooth framework is that it is able to naturally make use of the dynamically updated discrete disparities. This method is highly parallel to each pixel and suitable for GPU computation.

\section{Model Mosaicking}

\begin{figure*} [htp]
\vspace{0.0cm}
\centering
  \includegraphics[width=0.8\textwidth]{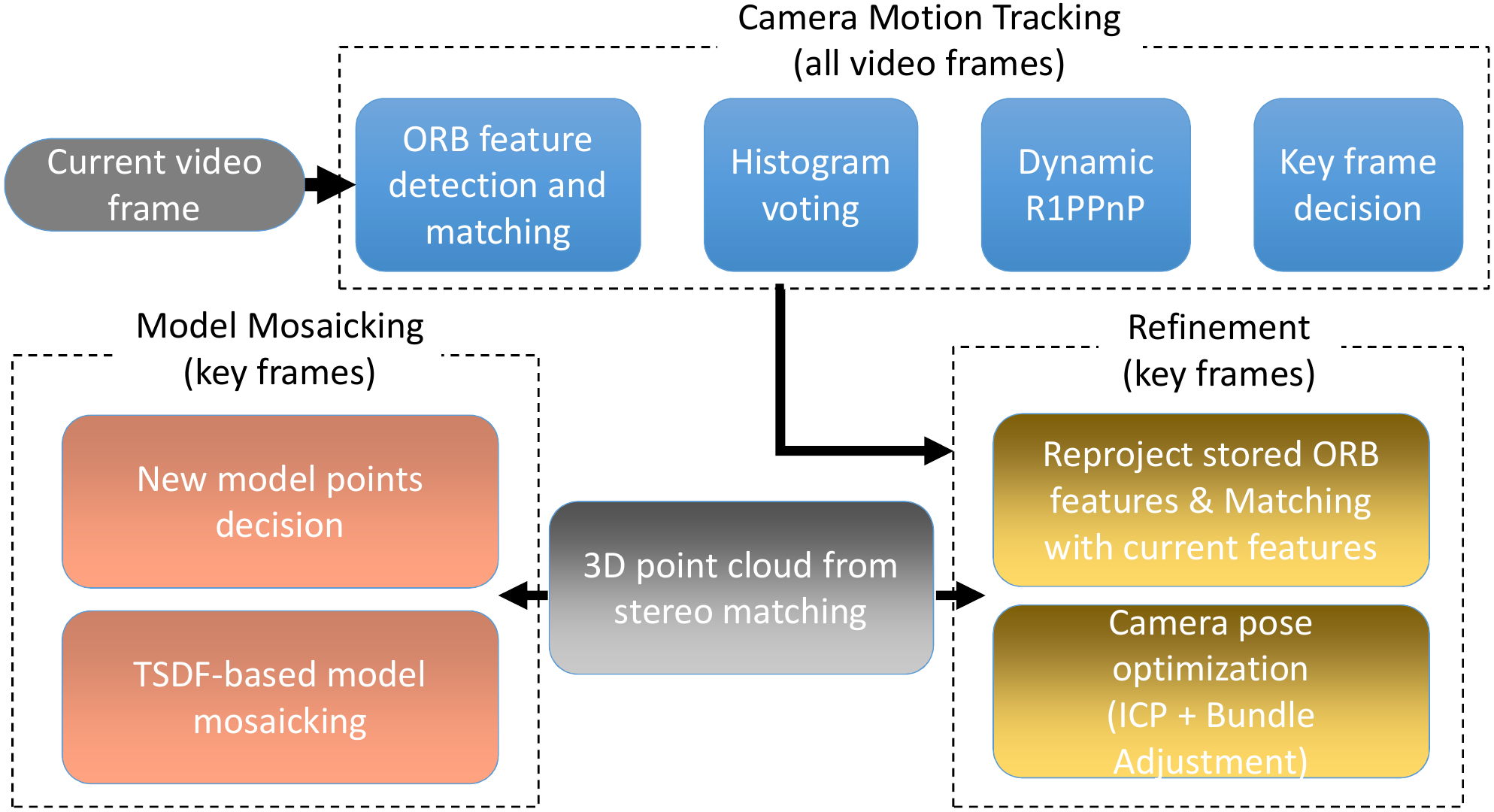}
  \caption{The flow chart of the SLAM method.}
\label{fig_slamprocess}
\end{figure*}

\begin{figure} [ht]
\vspace{0.0cm}
\centering
 \includegraphics[width = 0.45\textwidth]{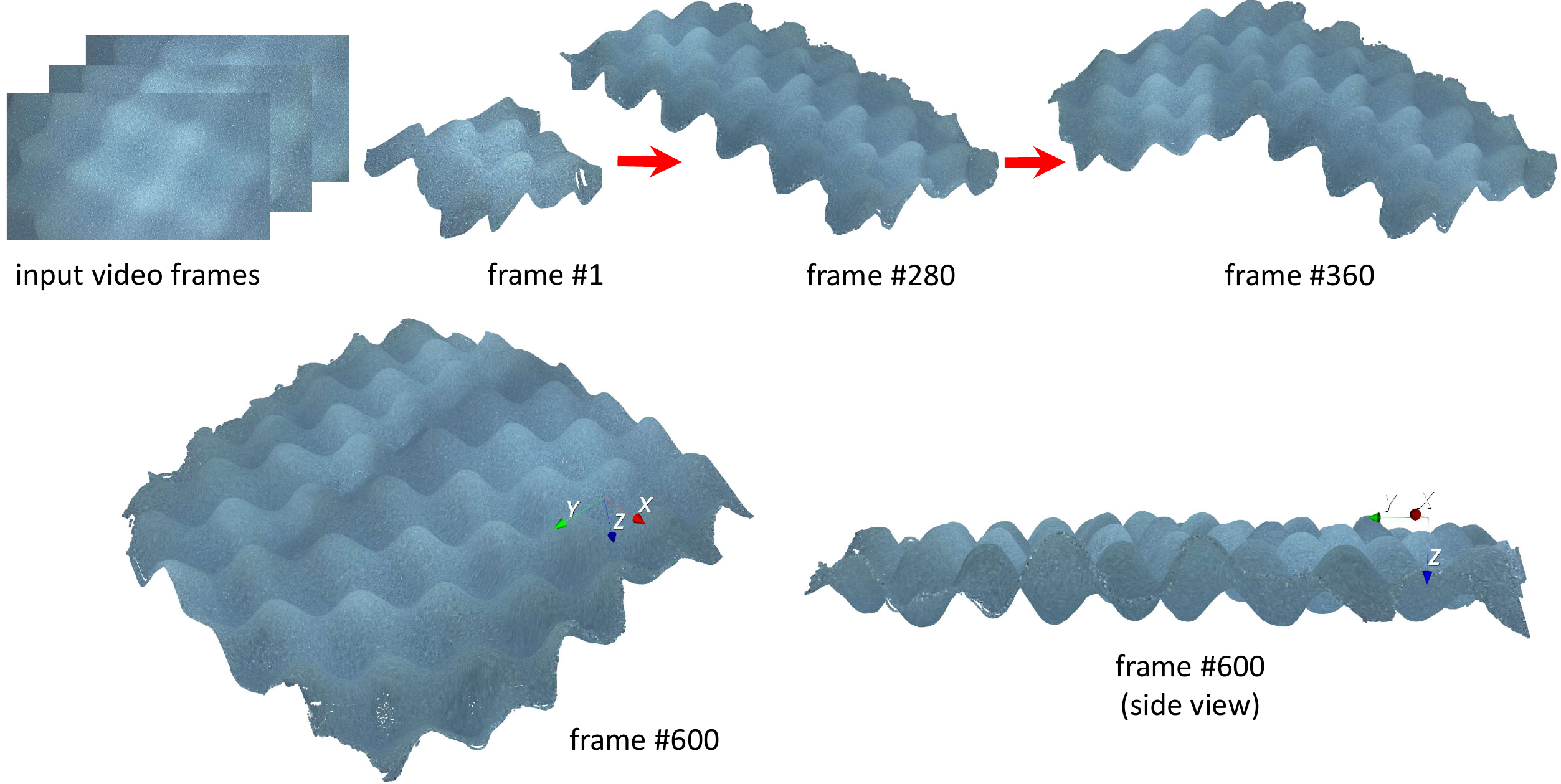}
 \caption{An example of our model mosaicking process with a phantom. }
\label{fig_spongeprocess}
\end{figure}

\begin{figure} [ht]
\vspace{0.0cm}
\centering
 \subfigure[]{
 \includegraphics[height=1.7cm]{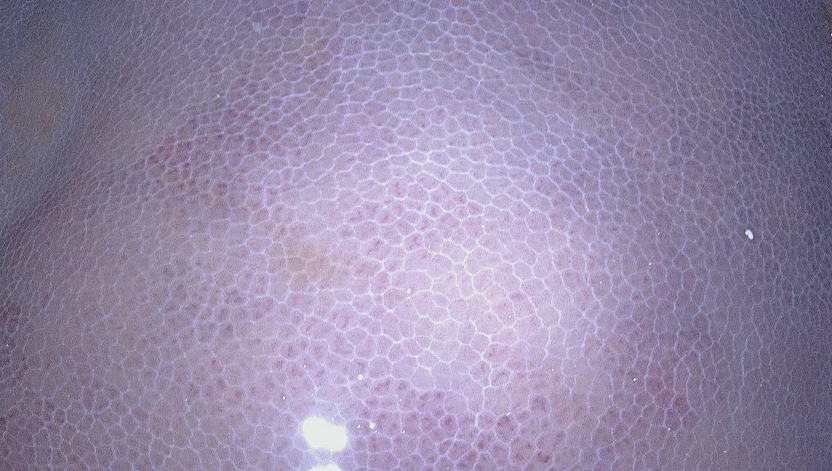}
}
 \subfigure[]{
 \includegraphics[height=1.7cm]{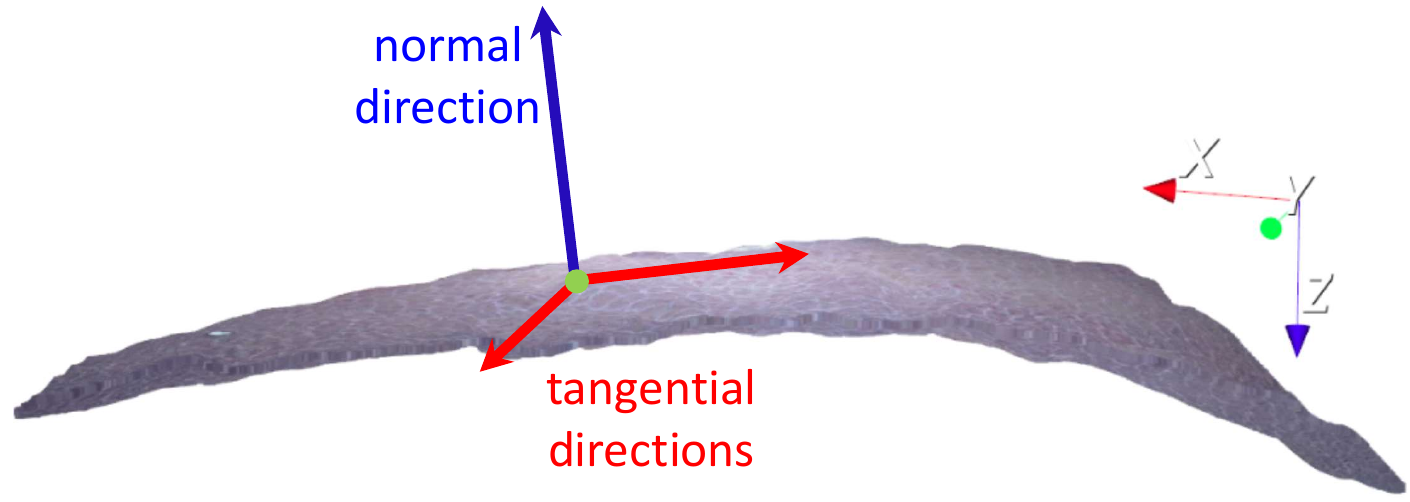}
}
 \caption{ICP-based model alignment may not work well due to the smooth tissue surface and the narrow field of view. We use the laparoscopies images of the liver surface as an example. To observe the texture clearly, the laparoscope should be close to the liver. (a) The obtained image has a narrow field of view. (b) The reconstructed 3D point cloud is small and smooth. Hence ICP-based alignment cannot find good constraints in the tangential directions, but is accurate in the normal direction.}
\label{fig_ICPdoesnotwork}
\end{figure}

We employ the truncated signed distance field (TSDF) method \cite{curless1996volumetric} to mosaic the raw 3D point cloud generated from pixel disparities results of stereo matching and the camera calibration parameters to obtain the extended 3D model of the tissue surface, as shown in Fig. \ref{fig_spongeprocess}. The prerequisite to perform TSDF is to align the raw 3D point cloud accurately, which is equivalent to the estimation of camera motion in this video-based 3D reconstruction problem. As shown in Fig. \ref{fig_ICPdoesnotwork}, conventional iterative closest points (ICP)-based model alignment is difficult to handle smooth tissue surfaces. Another way to align models is based on image feature points matching. However, due to the low texture and varying illumination condition, feature matching is challenging and a large amount of outliers may exist. To overcome these problems, we propose a novel SLAM method that consists of fast and robust algorithms to handle the large percentage of feature matching outliers in real-time.

The flow chart of our SLAM method is shown in Fig. \ref{fig_slamprocess}, which mainly consists of three modules. The first module tracks the camera motion between adjacent video frames according to ORB feature matching \cite{rublee2011orb}, which is mainly based on a novel and robust P$n$P algorithm called DynamicR1PP$n$P. The second module aims to refine the camera motion estimation results at key frames and eliminate the accumulative error, which is based on the minimization of ICP and bundle adjustment (BA) costs. The third module performs TSDF-based model mosaicking and manages feature points. In the following section, we will introduce the details of the involved algorithms.

\subsection{Histogram Voting-based Matching Inliers Preselection}

Our SLAM system is based on the ORB feature \cite{rublee2011orb}, which is much faster to detect and match than the conventional SURF feature \cite{bay2006surf}, and has been widely used in real-time SLAM systems, such as the ORB-SLAM method \cite{mur2017orb}.

However, the low and/or repeating texture of the tissue surface and varying illumination condition may result in a large amount of incorrect feature matches. In practice we observed that the percentage of outliers may be larger than $85\%$, making the traditional RANSAC+P3P\cite{kneip2011novel}-based outliers removal method slow. In addition, the small number of correct matches also decreases the accuracy of camera motion estimation. Hence, it is necessary to design algorithms to handle the large percentage of matching outliers.

ORB matching is performed between two adjacent video frames for camera motion tracking. Under a reasonable assumption that the camera motion, especially the roll angle, between adjacent video frames is minimal during the surface scan, we propose to utilize the displacements of matched ORB features between two adjacent images to roughly distinguish correct and incorrect ORB matches. Specifically, we denote the image coordinates of matched ORB features at two images as $[u_i^{(1)},v_i^{(1)}]$ and $[u_i^{(2)},v_i^{(2)}]$, $i=1,...,N$, where $u$ and $v$ are the $x-$ and $y-$ image coordinates in pixels. A correct match $k$ should have a similar displacement $[u_i^{(2)}-u_i^{(1)},v_i^{(2)}-v_i^{(1)}]$ with other correct matches. Hence, we first generate the histogram of $[u_i^{(2)}-u_i^{(1)},v_i^{(2)}-v_i^{(1)}]$, and then consider the ORB matches that are close to bins with large histogram value more likely to be inliers, which will be assigned with higher priority to be the control points for the subsequent DynamicR1PP$n$P algorithm. It should be clarified that this histogram voting-based inliers preselection step may not be 100\% correct, but it is fast and able to remove a large amount of outliers fast for the subsequent steps of the SLAM method.

\subsection{DynamicR1PP$n$P}

\begin{figure} [htp]
\vspace{0.0cm}
\centering
 \includegraphics[width=0.45\textwidth]{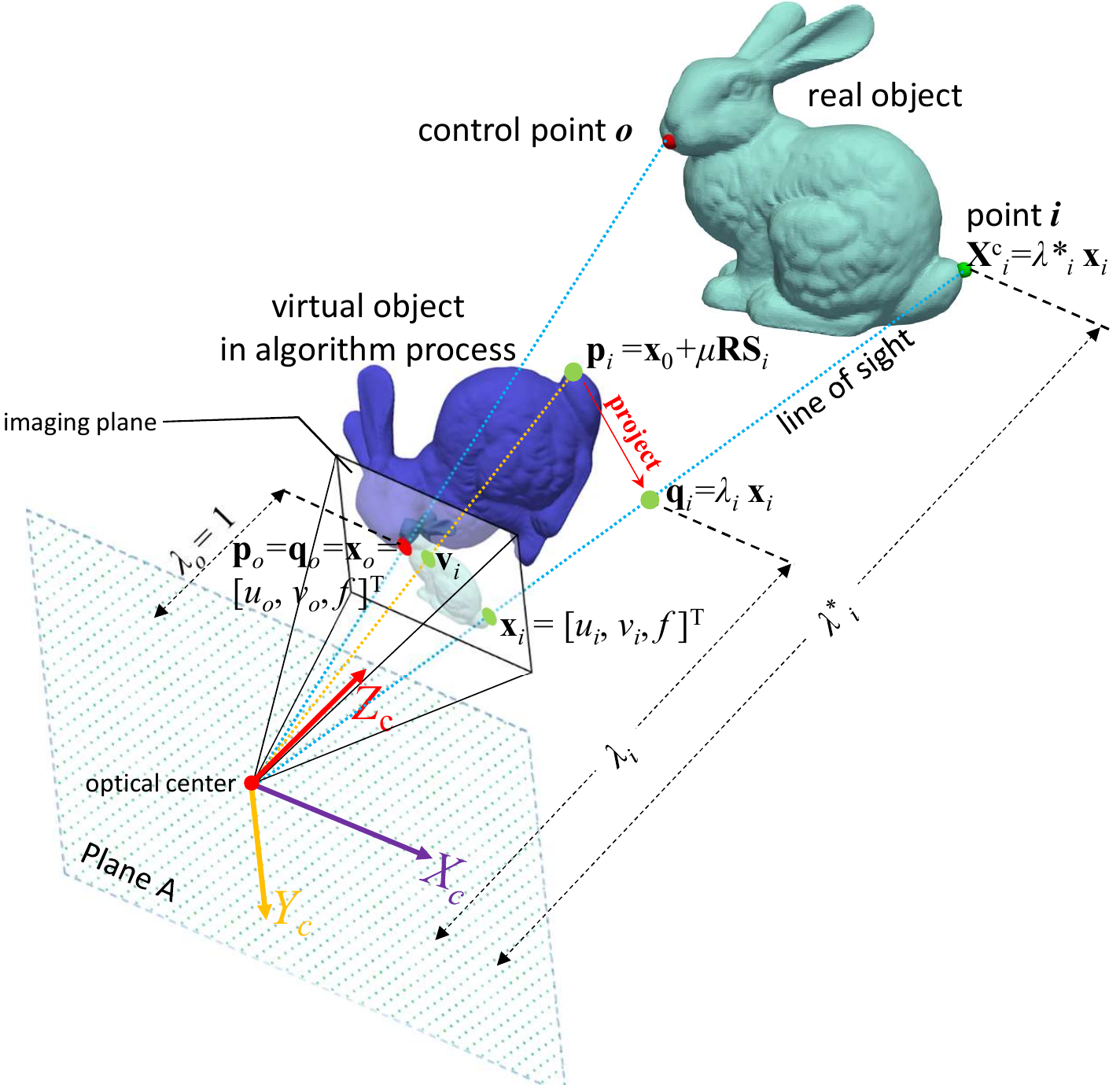}
 \caption{Demonstration of the geometrical relationships in the R1PP$n$P algorithm with a bunny model. The mouth point is used as the control point $o$ and the tail point is used to exemplify the geometrical relationships.}
\label{fig_bunny}
\end{figure}

\begin{figure} [htp]
\vspace{0.0cm}
\centering
 \includegraphics[width=0.35\textwidth]{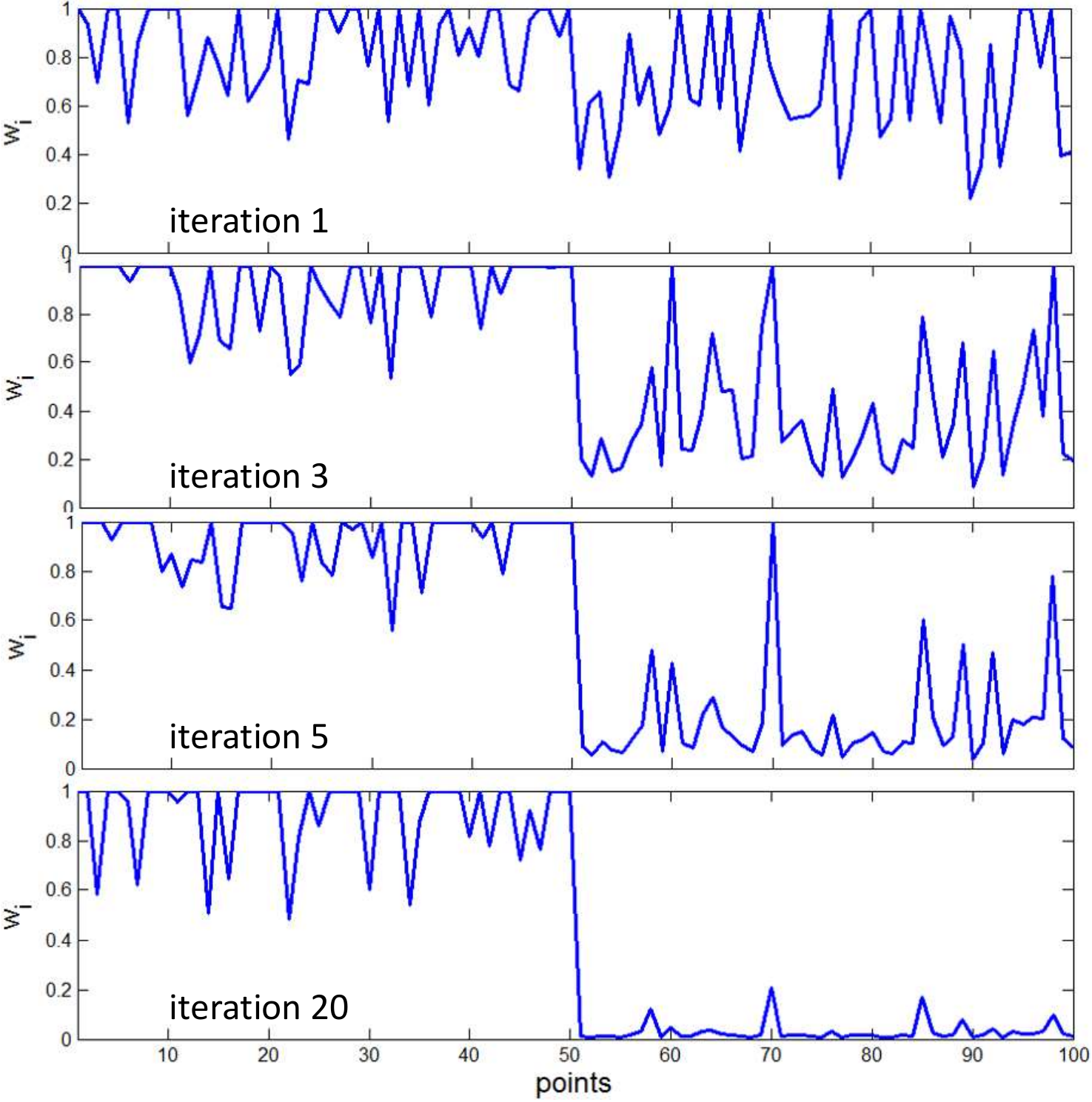}
 \caption{An example of the changes of weights $w_i$ in the R1PP$n$P iterations. In this example, the first 50 matches are inliers and the others are outliers. With the iteration, the weights of outliers decreases and their effects on camera motion estimation are reduced.}
\label{fig_r1ppnpweight}
\end{figure}

P$n$P methods, which aim to estimate the position and orientation of a calibrated camera from $n$ known matches between 3D object points and their 2D image projections, have been widely used in SLAM systems for camera motion estimation. We propose to modify and improve our previous R1PP$n$P work \cite{zhou2018re} to handle the problem of small number of matching inliers in the task of tissue surface reconstruction. In this section, we first briefly introduce the original version of R1PP$n$P and then introduce our modification.

R1PP$n$P is based on the standard pin-hole camera model, which is

\begin{equation}
{u_i} = f\frac{{x_i^c}}{{z_i^c}},{v_i} = f\frac{{y_i^c}}{{z_i^c}},
\label{eq_1}
\end{equation}

\noindent where $f$ is the camera focal length, ${\bf{x}}_i = {[{u_i},{v_i},f]^T}$ is the image homogeneous coordinate in pixels, and ${\bf{X}}_i^c = [x_i^c,y_i^c,z_i^c]^T$ is the real-world coordinate with respect to the camera frame. Hence, we have

\begin{equation}
{\bf{X}}_i^c = \lambda_i^* {\bf{x}}_i,
\label{eq_2}
\end{equation}

\noindent where $\lambda_i^* = z_i^c / f$ is the normalized depth of point $i$.

The relationship between the camera and world frame coordinate of point $i$ is
\begin{equation}
{\bf{X}}_i^c = {\bf{R}}{\bf{X}}_i^w + {\bf{t}},
\label{eq_3}
\end{equation}

\noindent where ${\bf{R}}\in SO(3)$ is the rotation matrix and ${\bf{t}}\in R^3$ is the translation vector. ${\bf{R}}$ and ${\bf{t}}$ are the variables that need to be estimated in the P$n$P problem. Selecting a point $o$ as the control point, we have

\begin{equation}
{\bf{X}}_i^c - {\bf{X}}_{o}^c = {\bf{R}}({\bf{X}}_i^w - {\bf{X}}_o^w),{\kern 1pt} {\kern 1pt} {\kern 1pt} {\kern 1pt} {\kern 1pt} {\kern 1pt} i \ne o.
\label{eq_4}
\end{equation}

Denoting ${\bf{S}}_i = {\bf{X}}_i^w - {\bf{X}}_o^w$, then, according to \eqref{eq_2} and \eqref{eq_4},

\begin{equation}
\lambda_i^* {\bf{x}}_i - \lambda_o^* {\bf{x}}_o = {\bf{R}}{\bf{S}}_i.
\label{eq_5}
\end{equation}

We divide both sides of \eqref{eq_5} by the depth of the control point $\lambda_o^*$, and rewrite \eqref{eq_5} as

\begin{equation}
\lambda_i {\bf{x}}_i - {\bf{x}}_o = \mu{\bf{R}}{\bf{S}}_i,
\label{eq_6}
\end{equation}

\noindent where $\lambda_i = \mu \lambda_i^*$ and $\mu=1/\lambda_o^*$ is the scale factor. We have

\begin{equation}
{\bf{t}} =  1/\mu {\bf{x}}_o - {\bf{RX}}_o^w.
\label{eq_7}
\end{equation}

\noindent which suggests that $\bf{t}$ can be computed from $\bf{R}$ and $\mu$.

The geometrical relationships of R1PP$n$P is shown in Fig. \ref{fig_bunny}. R1PP$n$P combines a re-weighting strategy and the 1-point RANSAC framework to reduce the effects of outliers. The 1-point RANSAC framework randomly selects one match as the control point $o$ and then alternatively update $\bf{R}$, $\mu$ and $\lambda_i$ to minimize the cost function

\begin{equation}
f({\bf{R}},\mu ,{\lambda _i}) = \sum\limits_{i = 1,i \ne o}^N {{w_i}{{\left\| {{\lambda _i}{{\bf{x}}_i} - {{\bf{x}}_o} - \mu {\bf{R}}{{\bf{S}}_i}} \right\|}^2}},
\label{eq_R1PPnPCost}
\end{equation}

\noindent where $w_i$ is the weight of point $i$ and is dynamically updated in the iteration process according to

\begin{equation}
\label{eq_reweighting}
{w_i} = \left\{ {\begin{array}{*{20}{c}}
{1.0}\\
{H/{e_i}}
\end{array}{\kern 1pt} {\kern 1pt} {\kern 1pt} {\kern 1pt} {\kern 1pt} {\kern 1pt} {\kern 1pt} {\kern 1pt} \begin{array}{*{20}{c}}
{if{\kern 1pt} {\kern 1pt} {e_i} \le H}\\
{if{\kern 1pt} {\kern 1pt} {e_i} > H}
\end{array}} \right.,
\end{equation}

\noindent where $e_i$ suggests the reprojection error of point $i$ with the current $\bf{R}$ and $\mu$ during iteration. $H$ is the inliers threshold that points with final reprojection errors smaller than $H$ are considered as inliers, and in our experiments we use $H = 5$ pixels. The reweighting rule \eqref{eq_reweighting} suggests that a point with a large reprojection error will have a small weight during the estimation of camera pose, as shown in Fig.\ref{fig_r1ppnpweight}. Our experimental results in Ref.\cite{zhou2018re} showed that R1PP$n$P has state-of-the-art performance compared with conventional RANSAC+P3P methods to handle matching outliers.

In our SLAM method, we will use the preselected matches according to the histogram voting results as the control points $o$ in the R1PP$n$P algorithm. However, when the number of feature matching inliers is small, R1PP$n$P or conventional RANSAC+P3P methods cannot estimate the camera motion accurately. Compared to obtaining correct matches, detecting consistent sets of feature points from two overlapped images is relatively easier. Based on this observation, we modified the R1PP$n$P method to dynamically update the feature matching relationships as follows: The camera pose is updated in an iterative process and the re-projection of stored feature points is updated accordingly. When the distance between a current feature point $i$ and the re-projected stored feature point $j$ is small, it should be considered as a possible correct match and we will add this candidate match dynamically, and the weight $w_{i,j}$ is updated according to

\begin{equation}
\begin{array}{l}
{w_{i,j}} := \begin{array}{*{20}{c}}
{\min \left( {H/{e_{i,j}},1} \right)}&{if{\kern 1pt} {\kern 1pt} {e_{i,j}} < \eta H}
\end{array}
\end{array},
\end{equation}

\noindent where $\eta > 1$ is a coefficient and in our experiments we use $\eta = 2.0$. Then, we perform normalization by

\begin{equation}
{w_{i,j}} := {w_{i,j}}/\left( {\sum\limits_j {{w_{i,j}}}  + {w_i}} \right),
\end{equation}

\noindent where $w_i$ is the weight of the original matches provided by ORB feature matching.

In order to eliminate incorrect matches that happen accidently in the iteration process, we decrease the weight $w_{i,j}$ if it is a newly observed candidate correspondence.

\begin{equation}
{w_{i,j}} := \min \left( {(k - {k_0})/T,1} \right){w_{i,j}}
\end{equation}

\noindent where $k$ is the current iteration index, $k_0$ is the first iteration index when this candidate correspondence is observed. $T$ is a pre-defined number and in our experiments we use $T = 5$.

\subsection{Key Frame Decision}

A frame is recognized as a key frame if it satisfies the following conditions: (1) There are at least 50 correct ORB matches when using DynamicR1PP$n$P, and (2) It has been at least 10 frames since the last key frame, or the difference between the camera pose of this frame and the last key frame is larger than a threshold. The camera pose difference is defined as

\begin{equation}
\left\| {{{\bf{t}}_{{\rm{diffence}}}}} \right\| + 20 \min \left( {\left\| {{{\bf{E}}_{{\rm{diffence}}}}} \right\|,2\pi  - \left\| {{{\bf{E}}_{{\rm{diffence}}}}} \right\|} \right),
\label{eq_postdiff}
\end{equation}

\noindent where ${\bf{E}}_{\rm{diffence}}$ and ${\bf{t}}_{\rm{diffence}}$ are the differences between the Euler angles and translations respectively. In our experiments we may use different pose thresholds for different data because the number of video frames and the tissue scale varies. A large pose threshold suggests that key frames are distant from each other and due to illumination changes, the textures on the mosaic may not look very smooth, but in general this pose threshold that determines key frames is not sensitive.

\subsection{Refinement of Camera Motion Estimation}

In the camera motion tracking stage, the 3D coordinates of feature points are directly obtained from stereo matching. The estimated camera poses are not accurate enough and bundle adjustment (BA)-based refinement \cite{triggs1999bundle} is necessary. We also take into account the ICP-based distance between current stereo matching model and the existing model to improve the robustness of the SLAM method because feature matching with previous key frames may fail due to low texture.

We first try to match the ORB features of current key frames with those of previous key frames to eliminate accumulative error. Our SLAM algorithm stores the feature points of previous key frames for reducing the accumulative error. With the camera motion tracking results, we select several previous key frames that have enough overlapped areas with the current key frame as the candidates. Then, we perform ORB matching with the candidate previous key frames and perform DynamicR1PP$n$P to detect correct matches.

Then we apply the optimization method to refine the camera motion estimation results. At a key frame with index $T$, we refine the camera pose estimation results by minimizing the cost function

\begin{equation}
{f_{{\rm{total}}}}({\bf{R}}_t,{\bf{t}}_t, {{\bf{x}}_i}) = {f_{{\rm{BA}}}}({\bf{R}}_t,{\bf{t}}_t, {{\bf{x}}_i}) + \beta {f_{{\rm{ICP}}}}({\bf{R}}_T,{\bf{t}}_T), t \in \Omega
\label{eq_totalcost}
\end{equation}

\noindent where $\Omega$ is the set of indices of video frames, which includes the current key frame $T$, all frames between the last key frame and current key frame $T$, and the matched previous key frames. ${\bf{R}}_t$ and ${\bf{t}}_t$ are the camera rotation and translation at video frame $t$ respectively. ${\bf{x}}_i$ is the coordinate of feature point $i$. $\beta$ is a weighting coefficient, which is dynamically adjusted according to the ratio of the number of feature points and ICP points. In our experiment we use $\beta  = {\rm{0}}{\rm{.1}} \times {\rm{number}}{\kern 1pt} {\kern 1pt} {\rm{of}}{\kern 1pt} {\kern 1pt} {\rm{feature}}{\kern 1pt} {\kern 1pt} {\rm{points}}/{\rm{number}}{\kern 1pt} {\kern 1pt} {\rm{of}}{\kern 1pt} {\kern 1pt} {\rm{ICP}}{\kern 1pt} {\kern 1pt} {\rm{points}}$.

The first term of \eqref{eq_totalcost}, ${f_{{\rm{BA}}}}( \cdot )$, is the standard local BA cost that aims to minimize the re-projection error, which only considers video frames that are included in $\Omega$. In this term, we fix the pose of the last key frame and the feature points observed in the last key frame to avoid scale drift.

The second term of \eqref{eq_totalcost}, ${f_{{\rm{ICP}}}}( \cdot )$, aims to minimize the distance between the existing 3D model and the current stereo matching model at key frame $T$, which is

\begin{equation}
{f_{{\rm{ICP}}}}({\bf{R}}_T,{\bf{t}}_T) = \sum\limits_i {\rho \psi ({{\bf{n}}_i}({\bf{R}}_T{{\bf{p}}_i} + {\bf{t}}_T - {{\bf{q}}_i}))},
\label{eq_ICPterm}
\end{equation}

\noindent where ${\bf{p}}_i$ are points of the existing model, and ${\bf{q}}_i$ are points of the current stereo matching model that has the same re-projection pixel coordinate with ${\bf{R}}_T{{\bf{p}}_i} + {\bf{t}}_T$. $\psi ( \cdot )$ is Tukey's penalty function to handle outliers. $\rho = 1$ if ${\bf{q}}_i$ has a valid depth, otherwise $\rho = 0$. ${\bf{n}}_i$ is the normal direction of ${\bf{q}}_i$ obtained from the stereo matching point cloud, which allows the template to 'slide' along the tangent directions, as shown in Fig. \ref{fig_ICPdoesnotwork}.

To minimize the cost function \eqref{eq_totalcost}, a GPU-based parallel Levenberg-Marquardt (LM) algorithm is developed. The equation in the standard LM algorithm to update the variables is

\begin{equation}
({{\bf{J}}^T}{\bf{J}} + \lambda diag({{\bf{J}}^T}{\bf{J}})){\bf{x}} = {{\bf{J}}^T}{\bf{b}},
\label{eq_jxb2}
\end{equation}

\noindent where $\bf{x}$ is the vector of variables, $\bf{J}$ is the Jacobian matrix and $\bf{b}$ is the residual vector. $\lambda$ is a parameter that controls the updating step.

According to Eq. \eqref{eq_totalcost}, the variables to be estimated consist of camera poses and coordinates of feature points. Since a feature point may exist in most of the recent video frames, the structure of the whole Jacobian matrix $\bf{J}$ is large and dense. In order to accelerate the computation, we split the variables into two parts in our LM implementation and update the two parts of variables alternatively.

To update the coordinates of feature points, because each feature point is independent with respect to each other when the camera poses are fixed, we launch one GPU thread for each feature point and calculate the related Jacobian matrix and the residual re-projection errors.

We update the camera poses of different frames separately. Because only the camera pose at key frames considers the ICP term \eqref{eq_ICPterm}, hence the main parameters of Eq. \eqref{eq_jxb2} at key frames can be split to

\begin{equation}
{{\bf{J}}^T}{\bf{J}} = {\bf{J}}_{{\rm{BA}}}^T{{\bf{J}}_{{\rm{BA}}}} + \beta^2 {\bf{J}}_{{\rm{ICP}}}^T{{\bf{J}}_{{\rm{ICP}}}},
\end{equation}

\noindent and

\begin{equation}
{{\bf{J}}^T}{\bf{b}} = {\bf{J}}_{{\rm{BA}}}^T{{\bf{b}}_{{\rm{BA}}}} + \beta^2 {\bf{J}}_{{\rm{ICP}}}^T{{\bf{b}}_{{\rm{ICP}}}}.
\end{equation}

We launch multiple parallel GPU threads to compute each row of $\bf{J}$. Then, we perform Cholesky decomposition to solve \eqref{eq_jxb2}.

\subsection{GPU-based TSDF Mosaicking}

The basic idea of TSDF is to take the average value of the 3D coordinates of an area if it is observed multiple times, which is more accurate than the results of a single observation. Raw 3D point cloud can be obtained from key video frames by using our stereo matching method. We incrementally mosaic the stereo matching results to generate the extended tissue surface model based on the camera motion estimation results of SLAM. The extended tissue surface models are also in the form of 3D point clouds. Because we aim to obtain high resolution textures to provide better surgical navigation, the extended surface model usually include millions of points and traditional volume-based TSDF method may take too large amount of computer memory. To avoid this problem, we store the 3D coordinate and the RGB color for each point in the GPU memory without using volume grids. To build correspondences between the extended surface model and the current stereo matching results, we project the extended surface model to the current imaging plane according to the camera pose estimation results. This rasterize process is performed by using GPU parallel computation and is fast. For each pixel with a valid depth value in the stereo matching results, the related point in the extended surface model is merged with the stereo matching results by using the TSDF method. Pixels that are not covered by the reprojection are considered as new points and will be added to the extended surface model.

Since the light source is often equipped at the tip of the imaging modality, hence the image edge is often darker than the central area. In order to generate smooth texture of the model, we also use TSDF-like merging to update the RGB color of points. During RGB color merging, the TSDF updating weight is 1.0 if the point is at the center area of the image, and decreases as it approaches the image edge.

\section{Experiments}

The source code was implemented in CUDA C++ and executed on a Dell desktop with an Intel Xeon X5472 3.00 GHz CPU and NIVIDA Titan X GPU. We used OpenCV to read the recorded videos and the results were visualized by the Visualization Toolkit (VTK). We collected \textit{ex-} and \textit{in vivo} stereo videos for the evaluation of our method, and details of the videos are provided in Tab. \ref{tab_params1} and \ref{tab_params2}, which includes the video length, number of frames, average tissue-camera distance, average camera motion speed, size of the tissue and the number of points of the reconstructed models.

\subsection{Ex Vivo Experiments}

\begin{table*}[htbp]
\caption{Parameters of Qualitative Experiments}
\centering
\begin{tabular}{c|c|c|c|c|c|c|c|c|c}
\hline
 & exvivo & exvivo & exvivo & exvivo & invivo& invivo& invivo&invivo& invivo\\
& stomach & phantom & liver1 & liver2 & neurosurgery& kidney& urethral&spine& Hamlyn\\
 &(Fig.8(a))&(Fig.8(b))&(Fig.8(c))&(Fig.8(d))&(Fig.10)&(Fig.11)&(Fig.12)&(Fig.13)&(Fig.14)\\  \hline
 video length (s) & 19.5 & 48.3 & 15.5 & 28.7 & - & 3.4 & 12.4 & 6.1 & 35 \\ \hline
 number of frames &293&725&232&430&5&86&186&91&961\\ \hline
  resolution (pixels)&960$\times$540&960$\times$540&960$\times$540&960$\times$540&720$\times $480&1024$\times$768&960$\times$540&960$\times$540&320$\times$240\\ \hline
average tissue-camera&&&&&&&&& \\
distance (mm)&87.3&129&67.2&71.6&76.0&92.0&97.7&95.3&49.9 \\ \hline
average speed (mm/s)&10.7&10.0&6.2&4.4& - & 17.2&6.4&27.3&3.2 \\ \hline
bounding box length (mm) &199 &	274 &137&164& $80$& $153$ & $161$ &$252$& $125$ \\ \hline
number of model points ($\times10^6$)&2.1&2.6&1.9&1.8&0.8&1.1&1.2&2.2&0.2 \\ \hline
key frames threshold (Eq.\eqref{eq_postdiff}):&10.0&10.0&10.0&10.0&1e-6&5.0&10.0&10.0&3.0 \\ \hline
\end{tabular}
\label{tab_params1}
\end{table*}

\begin{table*}[htbp]
\caption{Parameters of Quantitative Experiments (Fig.7)}
\centering
\begin{tabular}{c|c|c|c|c|c|c|c}
\hline
 &liver1&liver2&liver3&liver4&kidney1&kidney2&kidney3\\  \hline
 video length (s) &25.2&14.3&10.9&15.7&7.3&6.2&4.3\\ \hline
 number of frames&378&215&164&236&109&93&64\\ \hline
 resolution (pixels)&960$\times$540&960$\times$540&960$\times$540&960$\times$540&960$\times$540&960$\times$540&960$\times$540\\ \hline
average tissue-camera&&&&&&& \\
distance (mm)&107.9&82.7&77.1&173.5&84.5&85.9&87.8\\ \hline
average speed (mm/s)&3.5&8.5&6.4&4.7&8.6&5.2&6.9 \\ \hline
bounding box length (mm) &163&136&121&249&181&137&134\\ \hline
number of model points ($\times10^6$)&1.2&1.2&0.9&0.7&0.7&0.7&0.7 \\ \hline
key frames threshold (Eq.\eqref{eq_postdiff}):&10.0&10.0&10.0&10.0&10.0&10.0&10.0\\ \hline
\end{tabular}
\label{tab_params2}
\end{table*}

\begin{figure*} [htp]
\vspace{0.0cm}
\centering
 \subfigure[]{
 \includegraphics[height = 6.2cm]{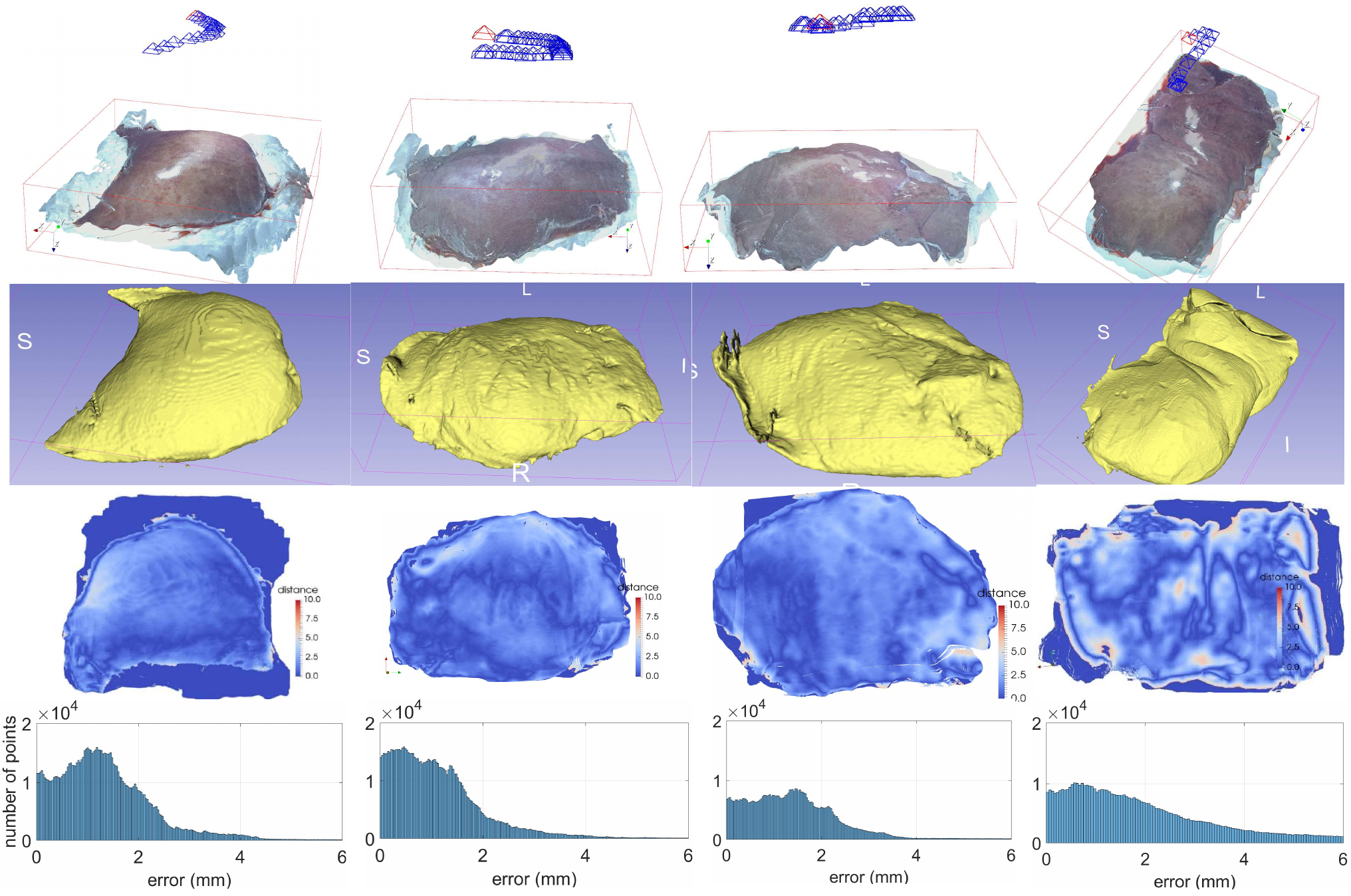}
}
 \subfigure[]{
 \includegraphics[height = 6.2cm]{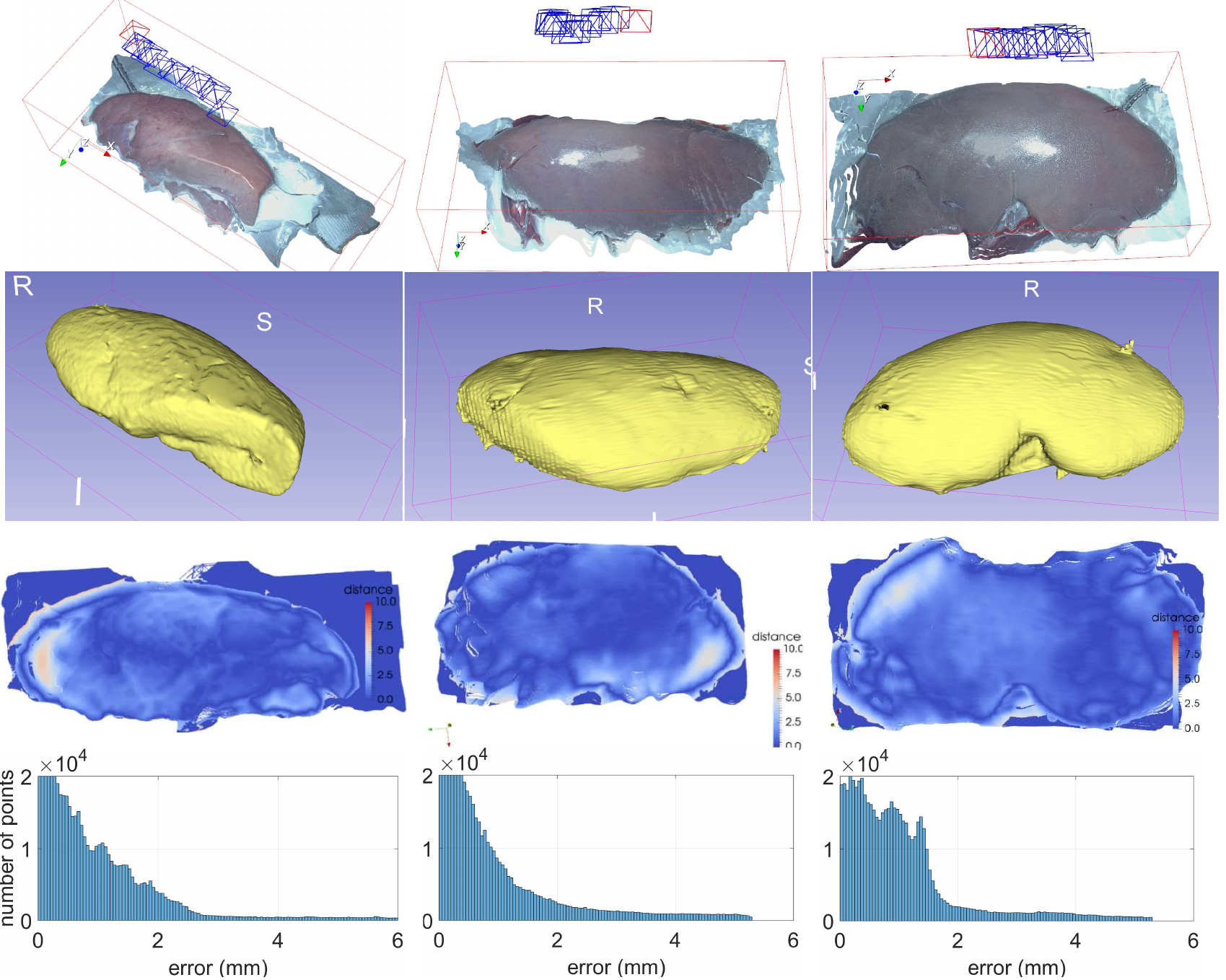}
}
 \caption{Quantified accuracy evaluation on stereo laparoscopy videos of ex-vivo porcine tissues. The 3D reconstruction results were compared with the CT segmentation results. The first row shows the 3D reconstruction results, the second row shows the CT image segmentation results, the third row shows the distance map after registration, and the last row shows the histogram of the errors. (a) Porcine livers, the RMSEs are 1.3, 1.1, 1.4 and 2.0 mm respectively. (b) Porcine kidneys, the RMSEs are 1.0, 1.0 and 1.1 mm respectively.}
\label{fig_quatifedwithCT}
\end{figure*}

\begin{figure*} [htp]
\vspace{0.0cm}
\centering
 \includegraphics[width=1.0\textwidth]{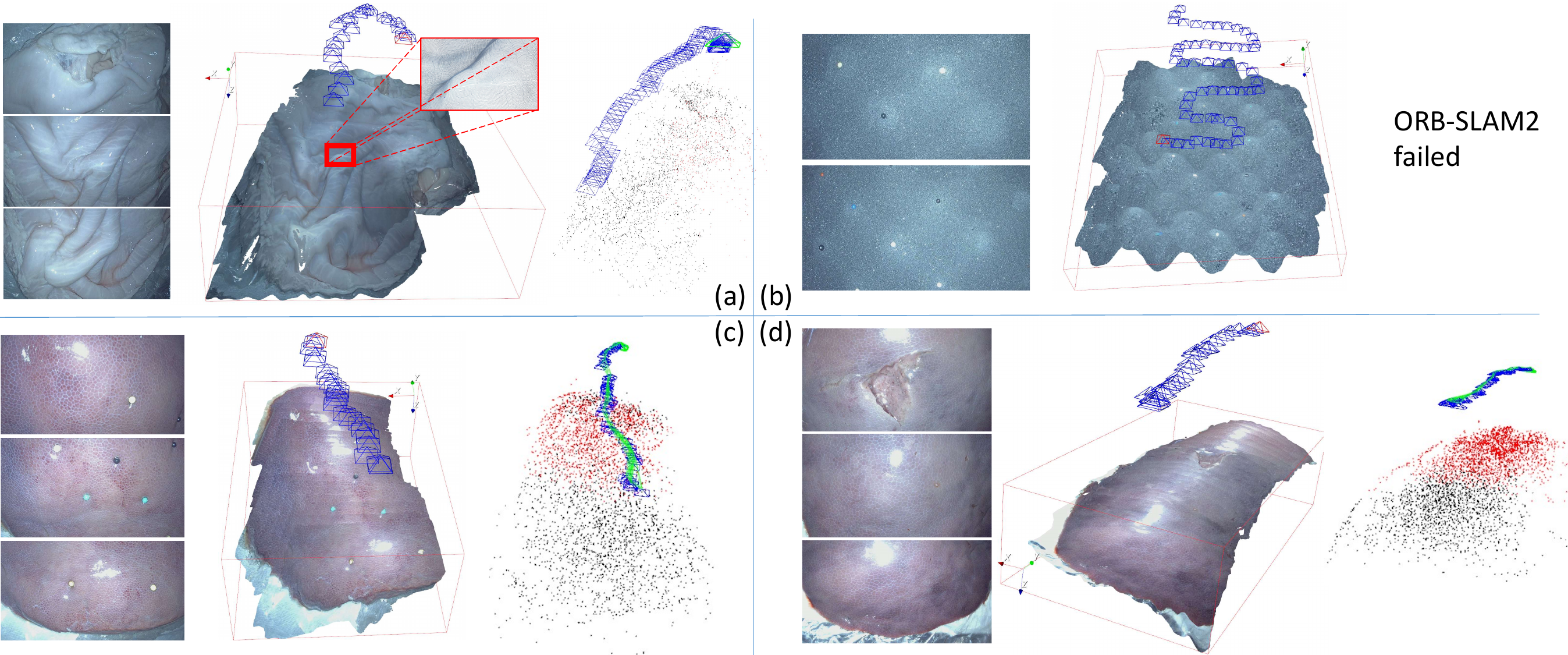}
 \caption{Qualitative results on stereo laparoscopy videos of phantoms and \textit{ex vivo} porcine tissues. The reconstructed tissues and the estimated camera motion (blue triangles) at key frames are shown in this figure. (a) A porcine stomach. (b) A phantom. (c)-(d) Porcine livers. A small region of the reconstructed model in (a) is enlarged to demonstrate the dense point cloud. For each case, from left to right are image samples (only images from the left camera are shown but both left and right images are used in our method), the reconstruction results of our method and the results of ORB-SLAM2. ORB-SLAM2 tracking failure occurred in cases in (b) and (d) due to the low texture.}
\label{fig_exvivo_all}
\end{figure*}

We first qualitatively tested our 3D reconstruction method on phantoms and \textit{ex vivo} tissues, including porcine stomachs and livers. We used a KARL-STORZ stereo laparoscope (model number TipCam 26605AA) with a resolution of $960 \times 540$ to capture stereo videos and performed the proposed 3D reconstruction method on the videos. The candidate disparity values for performing ZNCC matching are between $-20$ and $80$ pixels. Details of the videos are provided in Tab. \ref{tab_params1}. The results of our \textit{ex vivo} qualitative experiments are shown in Fig. \ref{fig_exvivo_all}. Since down-sampling is not included in the reconstruction process, the obtained 3D models have the same resolution as the input image, which usually include millions of points and are able to provide rich details of the surface texture. Our results qualitatively look promising and accurate. We also employed ORB-SLAM2 \cite{mur2017orb} for comparison, which is one of the most famous open-source SLAM methods. In order to handle low texture, the key parameters of ORB-SLAM2 were set as follows: the number of feature points is 3000 per image, and the threshold for detecting FAST corner is 1. As shown in Fig. \ref{fig_exvivo_all}(a) and (c), ORB-SLAM2 succeeded in reconstructing the sparse environment and tracking the camera motion. However ORB-SLAM2 tracking lost in cases shown in Fig. \ref{fig_exvivo_all}(b) and (d) due to the low texture.

\begin{figure} [ht]
\vspace{0.0cm}
\centering
 \subfigure[]{
 \includegraphics[height=3.5cm]{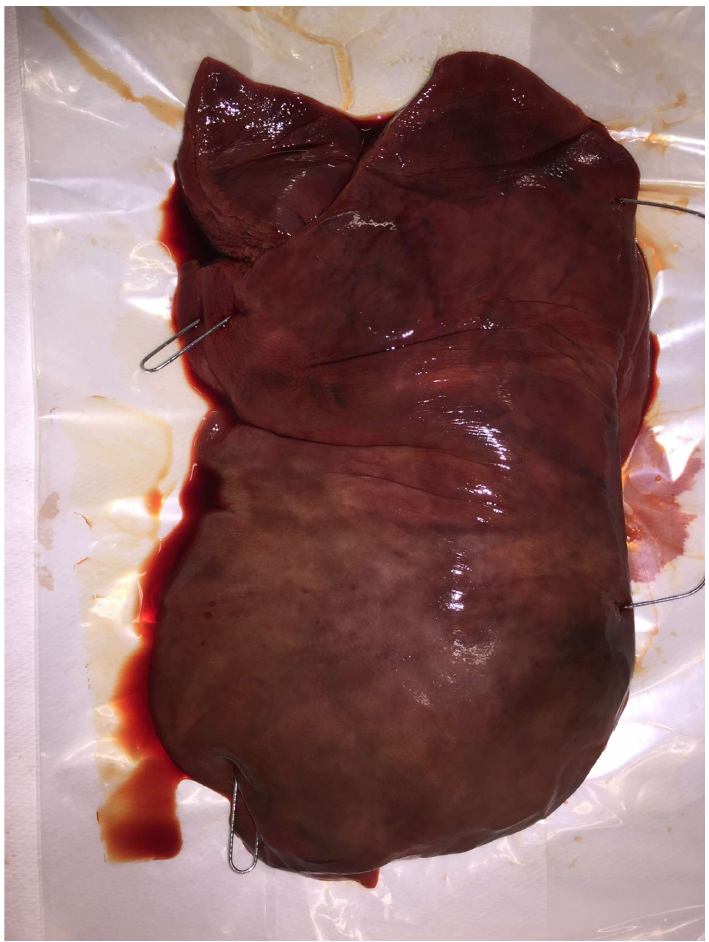}
}
 \subfigure[]{
 \includegraphics[height=3.5cm]{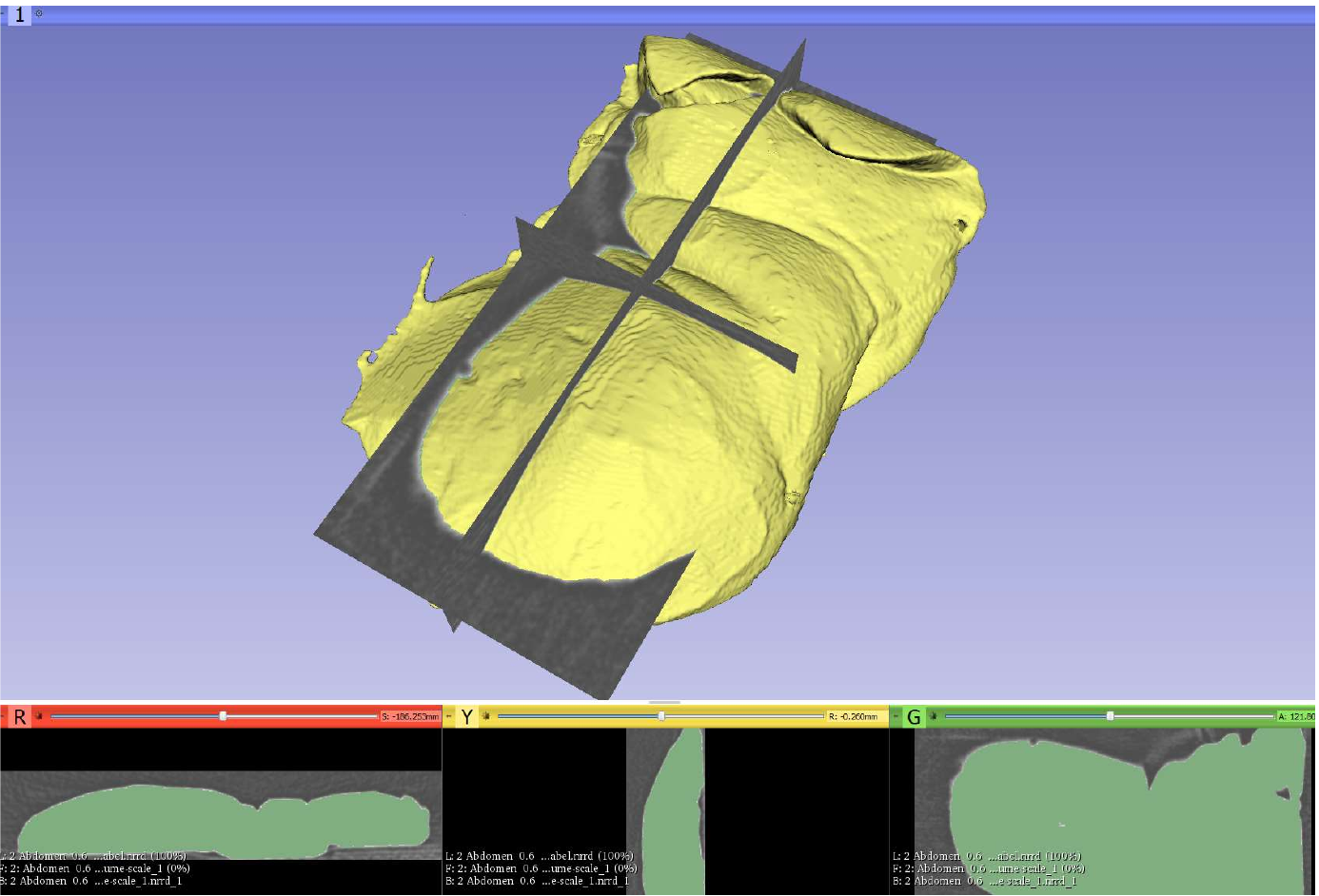}
}
 \caption{We used CT imaging of ex-vivo porcine tissues for quantified evaluation. (a) An ex-vivo porcine liver. (b) 3D Slicer-based segmentation of the obtained CT imaging.}
\label{fig_liverslicer}
\end{figure}

In order to evaluate the quantified accuracy of our 3D reconstruction method, we used the CT imaging of tissues as the gold standard. In this experiment, CT scans of four ex-vivo porcine livers and three kidneys were obtained (Siemens Somatom, Erlangen Germany) with a 0.6 mm resolution at our hospital, and we used the 3D Slicer software to segment the tissue models from the CT images, as shown in Fig. \ref{fig_liverslicer}. We captured stereo videos of the tissues with the KARL-STORZ stereo laparoscope, the details of which are in Tab. \ref{tab_params2}. Surfaces of livers and kidneys are very smooth and have low textures, but the proposed method was still able to reconstruct the 3D models, as shown in Fig. \ref{fig_quatifedwithCT}. To quantify accuracy, we registered the 3D reconstructed model with the CT segmentation results by first manually selecting landmarks, such as tissue tips, edge points and other recognizable points, and then refining the registration with the ICP algorithm. As shown in Fig. \ref{fig_quatifedwithCT} (a), the root mean square errors (RMSE) with the liver cases are 1.3, 1.1, 1.4 and 2.0 mm respectively. The fourth liver case has a relatively larger error because we used an entire piece of liver and the video was captured at a larger camera-tissue distance. The results on porcine kidneys are shown in Fig. \ref{fig_quatifedwithCT} (b), the RMSE of which are 1.0, 1.0 and 1.1 mm respectively. The histograms of errors are also provided in Fig. \ref{fig_quatifedwithCT}, which show that most points have an error of less than 2mm. It is worth noting that there are multiple sources of errors, including 3D reconstruction error, CT resolution error, CT segmentation error and registration error that contribute to the obtained RMSE in this experiment. In addition, because the livers and kidneys were placed on a textureless plastic sheet and part of the sheet were also included in the 3D reconstructed model, which is difficult to be totally removed (see the tissue edges in the distance maps of Fig. \ref{fig_quatifedwithCT}), so the quantified error may also include a small amount of the background. Therefore, it is a reasonable assumption that the actual error of our 3D reconstruction method is smaller than the reported RMSE.

\subsection{In Vivo Experiments}

To further evaluate the performance of our surface reconstruction method in real-world surgical scenarios, we obtained intraoperative videos from various stereo imaging modalities during surgeries performed in our hospital and online videos. The details of the videos are provided in Tab. \ref{tab_params1}. The videos were captured under an Institution Review Board approved protocol. Patient consent was waived since the analysis was performed retrospectively and no clinical decisions were affected.

\begin{figure} [htp]
\vspace{0.0cm}
\centering
 \includegraphics[width=0.50\textwidth]{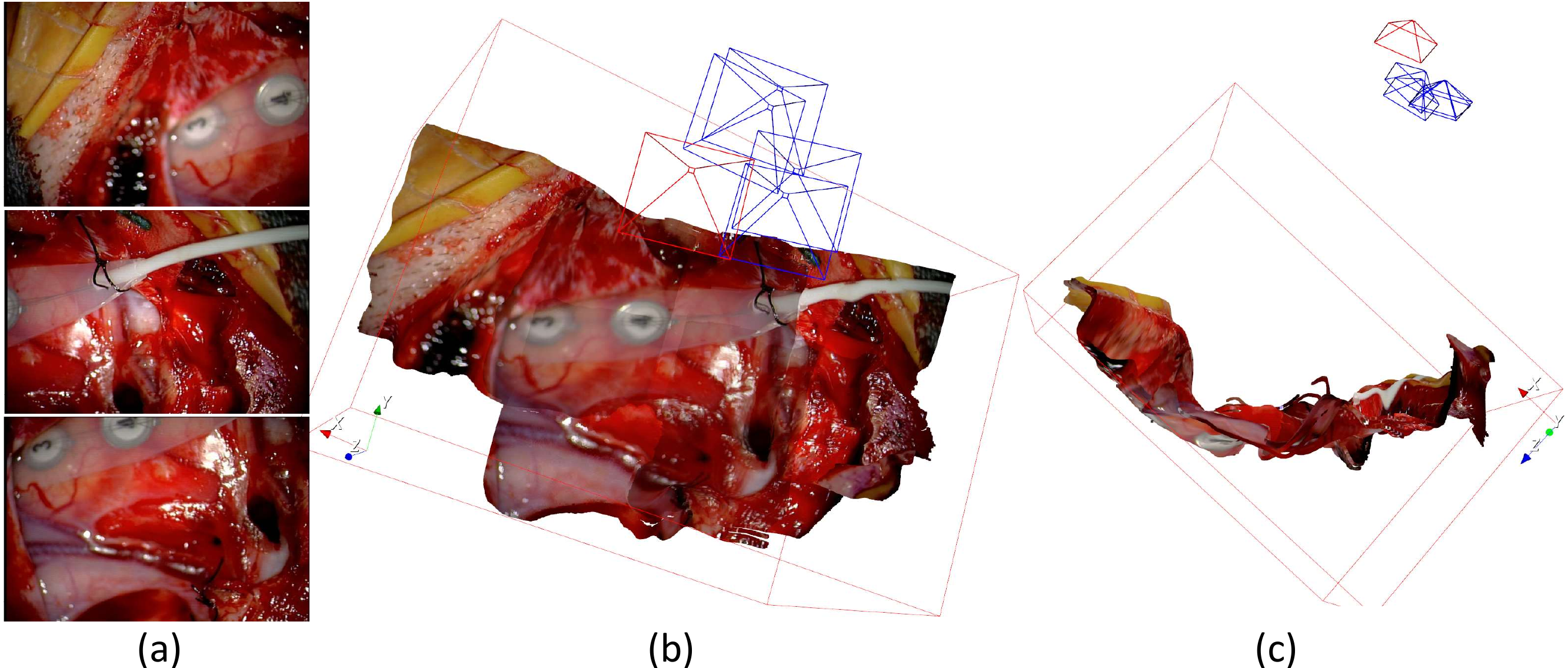}
 \caption{Experiments on stereo microscopic images captured during a neurosurgery at our hospital. (a) Samples of input images, and only images from the left camera are shown. (b)-(c) Our results.}
\label{fig_neuro_3D_reconstruction}
\end{figure}

\begin{figure} [htp]
\vspace{0.0cm}
\centering
 \includegraphics[width=0.50\textwidth]{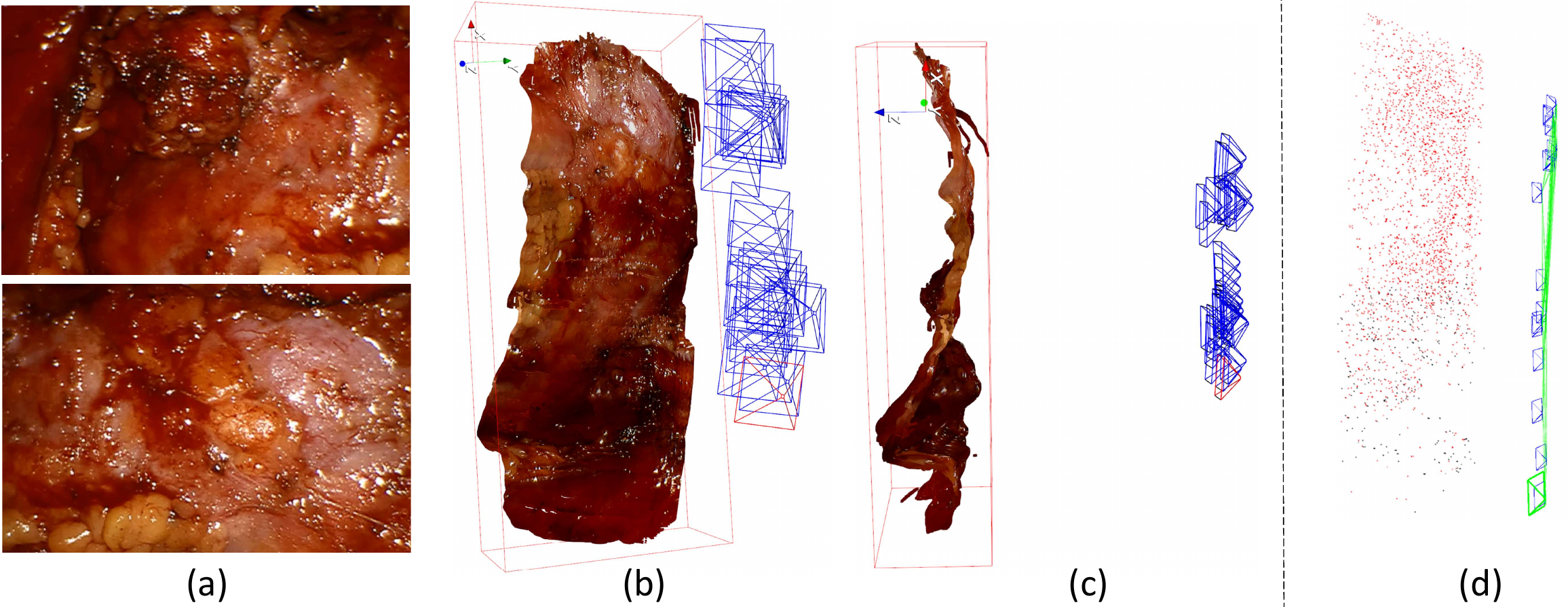}
 \caption{Experiments on stereo laparoscopy videos captured during a robotic kidney surgery at our hospital. The kidney surface and the tumor are shown in the images. (a) Samples of input left camera images. (b)-(c) Our results. (d) ORB-SLAM2 results. Due to respiration, the camera motion with respect to the kidney is more complex.}
\label{fig_invivo_kidney}
\end{figure}

\begin{figure} [htp]
\vspace{0.0cm}
\centering
 \includegraphics[width=0.50\textwidth]{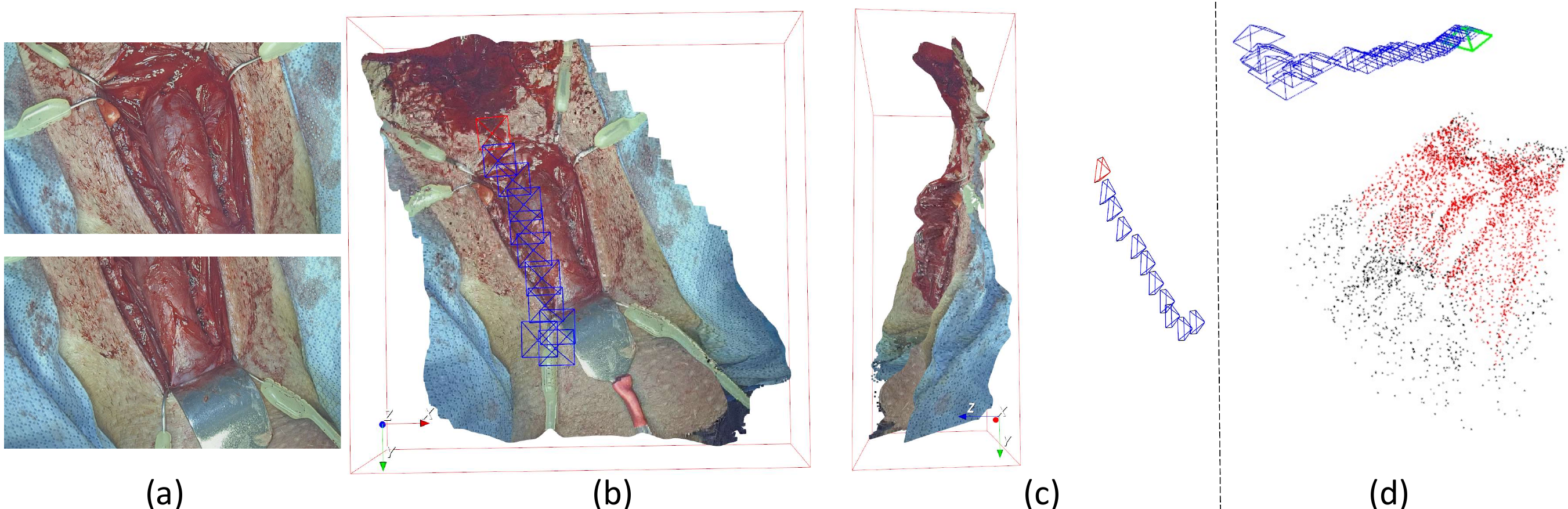}
 \caption{Reconstruction results of the urethra. We scanned the structures with the KARL STORZ stereo laparoscope during the surgery. (a) Samples of input left camera images. (b)-(c) Our results. (d) ORB-SLAM2 results.}
\label{fig_invivo_Urethra}
\end{figure}

\begin{figure} [htp]
\vspace{0.0cm}
\centering
 \includegraphics[width=0.50\textwidth]{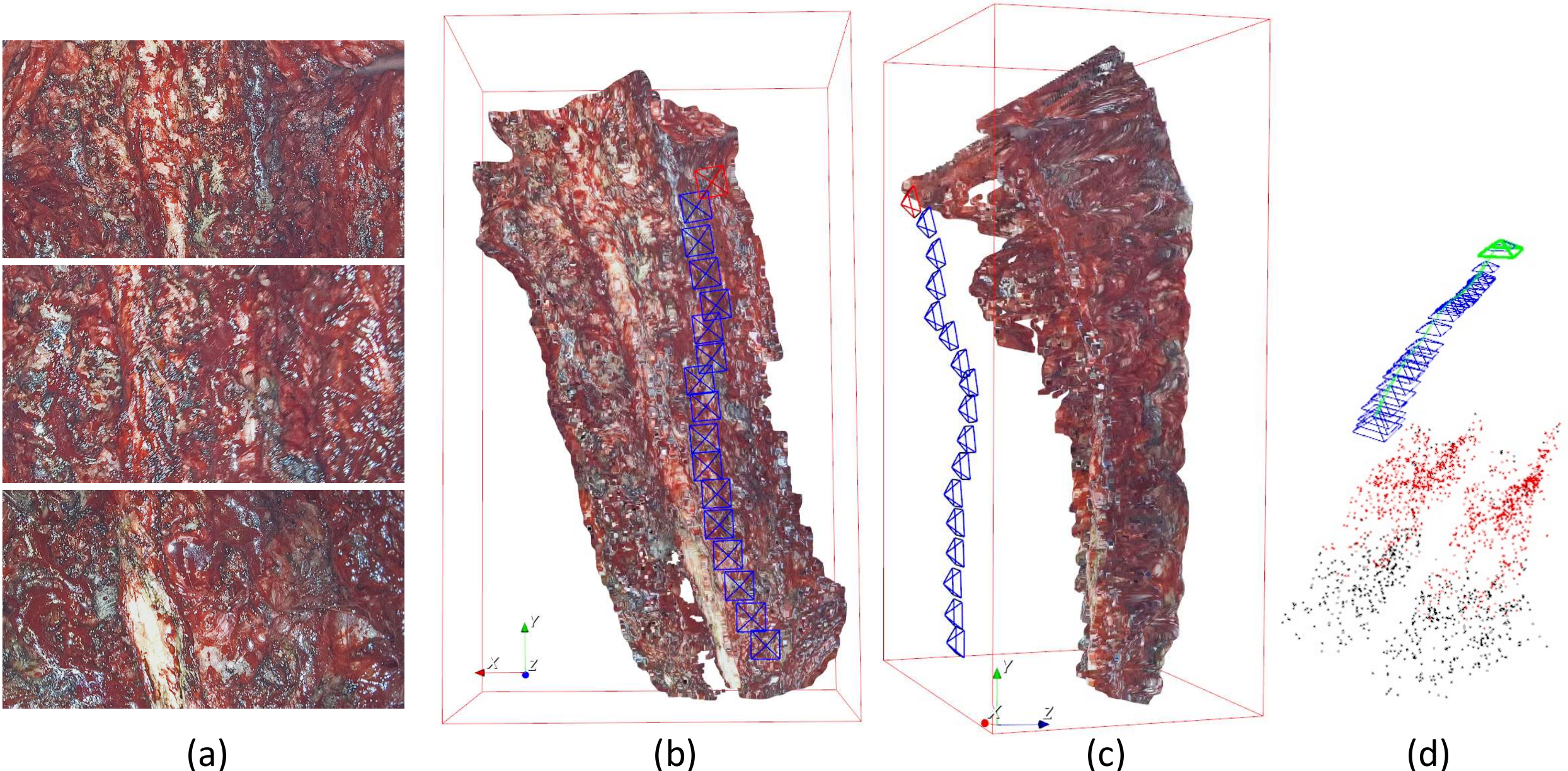}
 \caption{Reconstruction results of the spine. We scanned the structures with the KARL STORZ stereo laparoscope during the surgery. (a) Samples of input left camera images. (b)-(c) Our results. (d) ORB-SLAM2 results (tracking failed).}
\label{fig_invivo_Spine}
\end{figure}

\begin{figure} [htp]
\vspace{0.0cm}
\centering
 \includegraphics[width=0.45\textwidth]{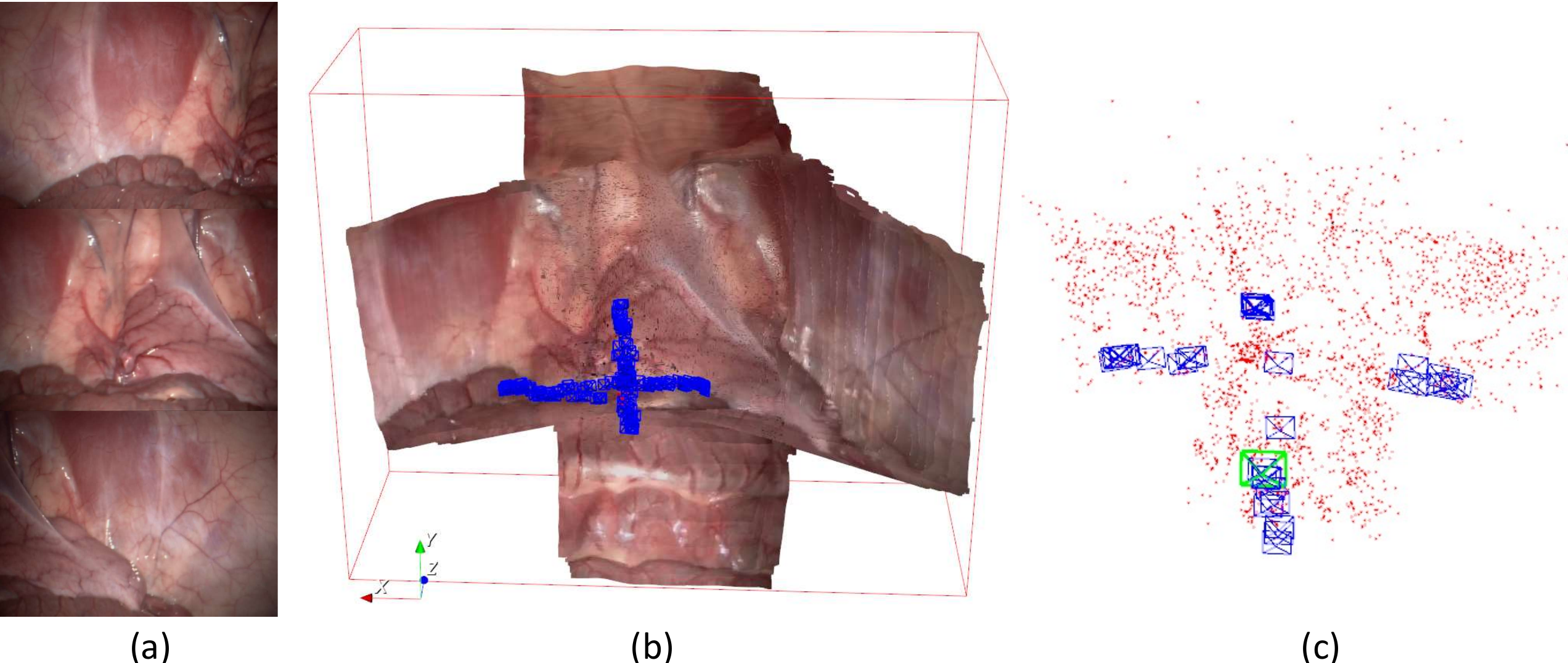}
 \caption{Experiments on in-vivo porcine abdomen videos from the Hamlyn datasets. (a) Samples of input left camera images. (b) Our results. (c) ORB-SLAM2 results.}
\label{fig_invivo_porcine}
\end{figure}

	For the first set of experiments, we obtained intraoperative stereo microscope images during a neurosurgery case. The dual channel output from a Carl-Zeiss microscope was captured using an Epiphan video capture card (DVI2PCI Duo) using the 3D Slicer software \cite{pieper20043d}. Five image frames with resolution $720 \times 480$ with small overlap between the frames were used to create a high-resolution mosaic of the surgical cavity. The results of the stereo reconstruction and mosaicking algorithms are shown in Fig. \ref{fig_neuro_3D_reconstruction}. In this experiment, we simply set the pose threshold to determine key frames to a small number hence all five images were used as key frames. Such a high-resolution mosaicking of the neurosurgery cavity could conceivably be used to register the intraoperative or diagnostic MRI to the mosaicked stereo reconstruction of the surgical cavity to identify remnant brain tumor during surgery. Due to the too small number of images, we did not run ORB-SLAM2 for this case.

	For the next set of experiments, we obtained high resolution stereo laparoscopy images of the kidney during a robot-assisted partial nephrectomy case. The video was obtained from the dual channel DVI output of the master console of the Intuitive da Vinci Xi robot. The video has the resolution of $1024\times768$, and was captured using two Epiphan video capture cards (DVI2PCI Duo) and a simple video capture program implemented using OpenCV. Prior to tumor resection, the surgeon scanned the exposed kidney surface using a stereo laparoscope. The 3D reconstructed model of the kidney surface and the tumor is shown in Fig. \ref{fig_invivo_kidney}. This model could further be registered to the diagnostic CT or MRI to plan the extent of surgical resection intraoperatively. This experiments also showed that our method can handle tissue motion caused by respiration, which is because respiration often cause the entire tissue to move but the deformation is relatively minimal. Since the time to scan the tissue surface is short, the tissue motion may not significant.

In the third set of experiments, we obtained intraoperative stereo laparoscopy images from a uretheroplasty procedure. Prior to resecting the urethral constriction, the urethra was exposed to identify the extent of the constriction. Thereafter, the authors scanned the exposed surgical area using a stereo laparoscope (Karl Storz Inc., model TipCam 26605AA) by moving the laparoscope slowly along the urethra. The interlaced video was captured and recorded using a video capture program in OpenCV. Fig. \ref{fig_invivo_Urethra} shows the results of the surface mosaicking algorithms of the exposed urethra. The figure shows a high-resolution 3D mosaicked surface model of the urethra and the surrounding structures. The fourth set of experiments were conducted with the same stereo laparoscope and the data was collected during a spine surgery, as shown in Fig. \ref{fig_invivo_Spine}. The spine bone was scanned by the Karl Storz stereo laparoscope after it was exposed. The estimated camera trajectories are smooth, which qualitatively prove that our method is accurate.

As shown in Fig. \ref{fig_invivo_porcine}, the last in-vivo experiment was conducted on the Hamlyn data \footnote{http://hamlyn.doc.ic.ac.uk/vision/data/Dataset1/stereo.avi}, which was captured within a porcine abdomen by using a stereo laparoscope. The length of this video is longer ($\approx 35$s) than other videos, and the smooth camera trajectory shown in Fig. \ref{fig_invivo_porcine} demonstrated that our method is able to work on such relatively long videos.

We also tested ORB-SLAM2 on the collected \textit{in vivo} data, which performed well on most cases because the texture is generally richer than our \textit{ex vivo} data. However, in the spine experiment we observed that ORB-SLAM2 failed to track the camera motion during the scan (see Fig. \ref{fig_invivo_Spine}).

Experiments with in-vivo data demonstrated that our approach can be applied to stereo optical videos obtained from different types of imaging modalities, and has potential for the 3D reconstruction of different types of tissues in varying lighting and surgical conditions. The reconstructed surface could be used for further registration to diagnostic or intraprocedural volumetric CT/MRI imaging.

\subsection{Runtime}

We report the average runtime of the main steps of the proposed 3D reconstruction method in Tab. \ref{tab_runtime}, which is the average results of 1,000 key frames on $960\times 540$ laparoscopy videos. The average computational time to process a key frame is 76.3 ms, which suggests that the proposed method is real-time.

\begin{table}[htbp]
\caption{Average Runtime of 3D Reconstruction(ms)}
\centering
\begin{tabular}{c|c}
\hline
video reading        & 12.7 \\ \hline
stereo matching      & 15.9 \\ \hline
ORB matching and histogram voting         & 15.2 \\ \hline
DynamicR1PPnP        & 6.1  \\ \hline
Refinement           & 24.6  \\ \hline
TSDF  & 2.2  \\ \hline
\textbf{Total}                & \textbf{76.3} \\ \hline
\end{tabular}
\label{tab_runtime}
\end{table}

\section{Conclusions}

In this paper, we have proposed a series of algorithms to solve the problem of tissue surface reconstruction, and mainly addressed the difficulties caused by low texture. The main novelties of this paper are as follows: (1) We have proposed effective post-processing steps for the local stereo matching method to enlarge the radius of constraint, and these steps are appropriate for GPU computation. (2) We have combined a histogram voting-based inliers pre-selection method and a novel DynamicR1PP$n$P algorithm that is robust to feature matching outliers to handle the camera motion tracking problem in the SLAM system. Traditional SLAM systems, such as ORB-SLAM, usually utilize RANSAC + P3P methods for camera motion tracking, which cannot work robustly when the number of inliers is too small. The methods proposed in this paper can greatly improve the robustness of the SLAM system. Experimental results on \textit{ex-} and \textit{in vivo} videos captured using different types of imaging modalities have demonstrated the feasibility of our methods, and the obtained models have high quality textures and the same resolution as the input videos. We have also introduced the CUDA implementation details to accelerate the computation with the GPU and enable real-time performance.

One limitation of this work is that we assume a static environment during the scan, hence this method is mainly suitable for surgeries on tissues with minimal deformation, such as the cases in our \textit{in vivo} experiments or other surgeries on bony structures. But such minimal deformation cases are common, which makes our method valuable for practical applications.

\section*{Acknowledgment}
This work was supported by the National Institute of Biomedical Imaging and Bioengineering of the National Institutes of Health through Grant Numbers R01EB025964, P41EB015898, P41RR019703, and a Research Grant from Siemens-Healthineers USA. We appreciate the generous help of Drs. Jiping Wang, Matthew Ingham, Steven Chang, Jairam Eswara, Carleton Eduardo Corrales, Sarah Frisken and Alexandra Golby in collecting \textit{in vivo} data.

\ifCLASSOPTIONcaptionsoff
  \newpage
\fi

\bibliographystyle{IEEEtran}
\bibliography{IEEEabrv,bare_jrnl}

%

\end{document}